\documentclass[sigconf]{acmart}

\setcopyright{acmlicensed}
\copyrightyear{2025}
\acmYear{2025}
\acmDOI{XXXXXXX.XXXXXXX}
\acmConference[Conference acronym 'XX]{Make sure to enter the correct
  conference title from your rights confirmation email}{June 03--05,
  2018}{Woodstock, NY}
\acmISBN{978-1-4503-XXXX-X/2018/06}

\usepackage{graphicx}
\usepackage{amsmath}
\usepackage{booktabs}
\usepackage{url}
\usepackage{multirow}
\usepackage{subcaption}
\usepackage{array}
\usepackage{balance}
\begin{document}
\title{DIGMAPPER: A Modular System for Automated Geologic Map Digitization [Industry]}

\author{Weiwei Duan, Michael P. Gerlek, Steven N. Minton}
\affiliation{%
  \institution{Inferlink Corporation}
  \city{El Segundo}
  \state{CA}
  \country{USA}
}
\email{{wduan, mgerlek, sminton}@inferlink.com}

\author{Craig A. Knoblock, Fandel Lin}
\affiliation{%
  \institution{USC Information Sciences Institute}
  \city{Marina del Rey}
  \state{CA}
  \country{USA}
}
\email{{knoblock, fandelli}@isi.edu}
\author{Theresa Chen, Leeje Jang, Sofia Kirsanova, Zekun Li, Yijun Lin,  Yao-Yi Chiang}
\affiliation{%
  \institution{University of Minnesota}
  \city{Minneapolis}
  \state{MN}
  \country{USA}
}
\email{{chen7924, jang0124, kirsa002}@umn.edu}
\email{{li002666, lin00786, yaoyi}@umn.edu}

\renewcommand{\shortauthors}{Weiwei Duan, et al.}

\begin{abstract}
Historical geologic maps contain rich geospatial information—such as rock units, faults, folds, and bedding planes—that is critical for assessing mineral resources essential to renewable energy, electric vehicles, and national security. However, digitizing maps remains a labor-intensive and time-consuming task. We present DIGMAPPER, a modular, scalable system developed in collaboration with the United States Geological Survey (USGS) to automate the digitization of geologic maps. DIGMAPPER features a fully dockerized, workflow-orchestrated architecture that integrates state-of-the-art deep learning models for map layout analysis, feature extraction, and georeferencing. To overcome challenges such as limited training data and complex visual content, our system employs innovative techniques, including in-context learning with large language models, synthetic data generation, and transformer-based models. Evaluations on over 100 annotated maps from the DARPA-USGS dataset demonstrate high accuracy across polygon, line, and point feature extraction, and reliable georeferencing performance. Deployed at USGS, DIGMAPPER significantly accelerates the creation of analysis-ready geospatial datasets, supporting national-scale critical mineral assessments and broader geoscientific applications.
\end{abstract}



\maketitle
\setlength\abovecaptionskip{3pt plus 0pt minus 0pt}
\setlength\belowcaptionskip{0pt plus 3pt minus 0pt}
\setlength{\textfloatsep}{0pt plus 1pt minus 0pt} 

\section{Introduction}
\begin{figure}[htbp]
\centering
\includegraphics[width=0.95\linewidth]{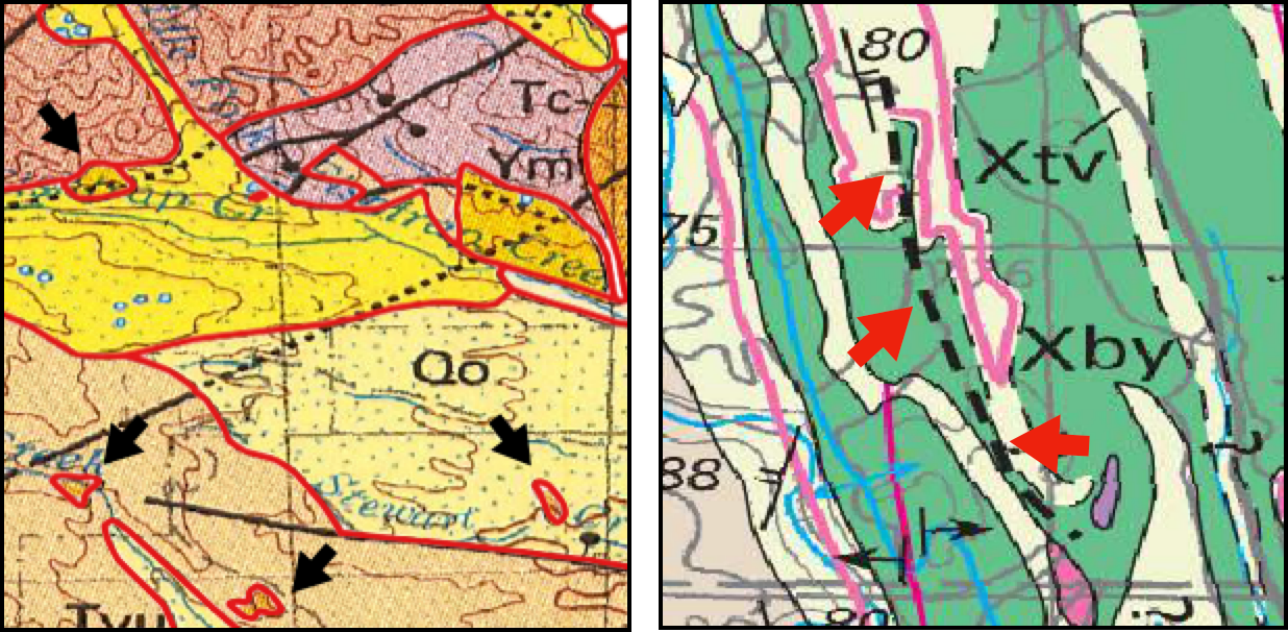}
\caption{Examples of complex visual context in geologic maps. Left: Polygon features with similar yellow tones are difficult to distinguish. The red lines delineate boundaries, and black arrows point to polygons visually similar to adjacent polygons. Right: Identifying a fault line (highlighted by red arrows) is challenging due to overlapping features, densely packed text, and surrounding visual clutter.}
\label{fig:challenge_example}
\end{figure}

The rising demand for critical minerals, such as lithium, cobalt, and rare earth elements, for renewable energy, electric vehicles, and advanced electronics, underscores the need for accurate geologic data. The United States Geological Survey (USGS) maintains an extensive collection of geologic maps that contain detailed information on various geologic features, including rock formations, faults, and mineral deposits. These geologic maps are valuable resources for critical mineral assessments~\cite{future_ml_mine,ml_mine,ai_mine}. However, manual digitization is a labor-intensive and time-consuming process~\cite{10.1145/2557423,10.5555/3383708,review_ml_mine,cnn_railroads}. In addition, automatic map digitization using deep learning faces two primary challenges: limited annotated training data and the complex visual map context. First, annotating geologic maps is laborious and requires expert knowledge, resulting in limited high-quality training data that hinder model performance~\cite{10.5555/3383708,label_correction}. Moreover, state-of-the-art (SOTA) models pre-trained on natural images do not work well on maps due to the significant differences in visual style and content~\cite{map_challenge_1}. Second, maps are densely populated with visually similar features, overlapping elements, and complex background patterns, making it difficult to accurately distinguish individual features. Figure~\ref{fig:challenge_example} (left) shows polygons with similar yellow tones, and (right) highlights fault lines amid visual clutter. The visual complexity also hampers georeferencing, as geologic maps often differ significantly in appearance from reference basemaps, making existing tools like ArcGIS~\cite{arcgis_georeferencing} ineffective. In summary, limited training data and complex visual context pose challenges to accurate map digitization. 

\setlength{\fboxsep}{1pt} 

\begin{figure*}[htbp]
    \centering
    \includegraphics[width=\linewidth]{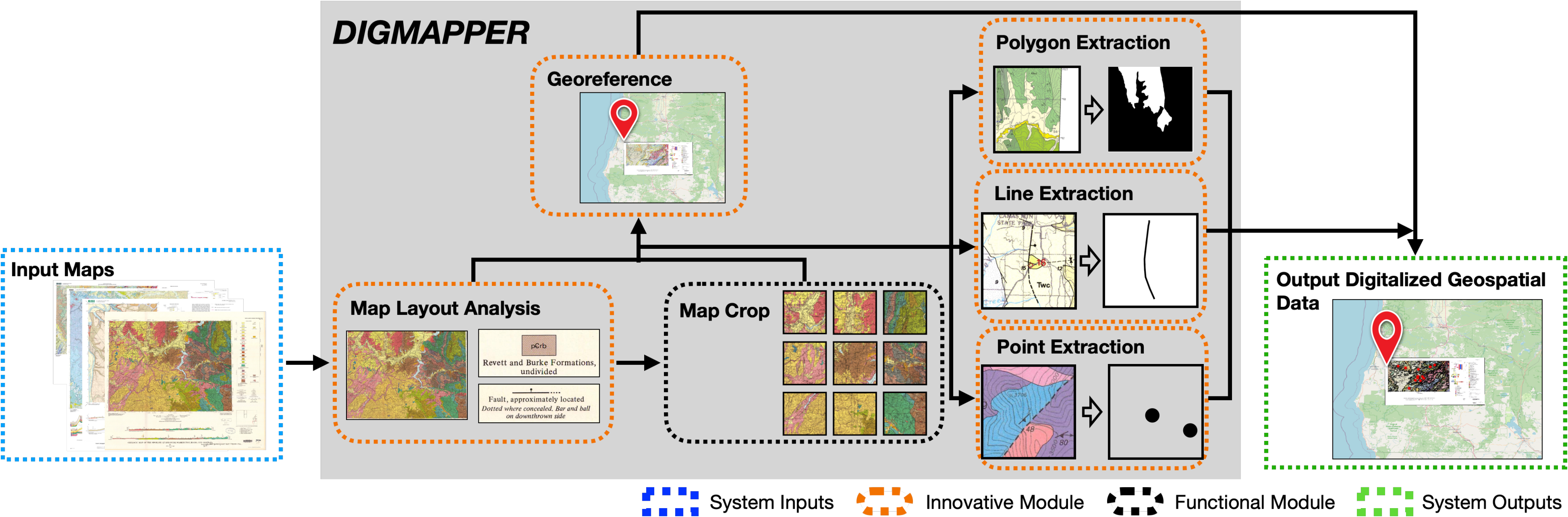}
    \caption{DIGMAPPER architecture for automated map digitization. The system takes a scanned map as input (blue) and processes it through a sequence of modular components. Innovative modules (orange) incorporate proposed deep learning models and training strategies for layout analysis, georeferencing, and the extraction of polygon, line, and point features. Functional (rule-based) modules (black), such as the map crop module, handle essential preprocessing tasks. The final output (green) includes a georeferenced map, as well as polygon, line, and point features.}
    \label{fig:sys_arch}
\end{figure*}

 To address these challenges and accelerate the identification of critical mineral locations, the US Defense Advanced Research Projects Agency (DARPA) and the USGS launched the Critical Mineral Assessments with Artificial Intelligence (AI) Support (CriticalMAAS) program. This paper presents our work under the CriticalMAAS program, specifically \textbf{DIGMAPPER}, illustrated in Figure~\ref{fig:sys_arch}, a modular system for automated map digitization. DIGMAPPER begins with Map Layout Analysis, segmenting the map into content areas and legend items (Section~\ref{sec:layout}), followed by a rule-based, functional Map Crop module that divides large map areas into smaller patches for parallel processing. The pipeline then branches into two parallel paths: one extracts map features using detected legend items, while the other performs georeferencing by aligning map content with real-world coordinates. To handle complex visual contexts in these tasks, we propose innovative deep learning mechanisms that enable accurate feature extraction and georeferencing (Sections~\ref{sec:polygon}, \ref{sec:line}, and~\ref{sec:georeference}). To address limited annotated data, we introduce methods for generating synthetic data to enhance the training data diversity (Section~\ref{sec:point}). Moreover, we explore in-context learning~\cite{incontext_survey} with Large Language Models (LLMs) to mitigate the need for extensive manual annotations (Section~\ref{sec:layout}). The final outputs are structured, analysis-ready geospatial data, including georeferenced maps and extracted features enriched with semantic labels derived from map legends. DIGMAPPER implements each module as an independent Docker image, coordinated by an orchestration layer that manages execution flow and supports seamless module integration or replacement (Section~\ref{sec:system}). This architecture ensures scalability, adaptability, and robustness for large-scale and long-term map digitization workflows.

DIGMAPPER has been successfully transitioned to the USGS to support automatic digital mapping workflows. Beyond USGS, DIGMAPPER also benefits a broader user community, including agencies involved in natural resource management, researchers studying Earth sciences, and private-sector users such as mining companies and geospatial data providers. 

\section{DIGMAPPER Modules and Performance}\label{method}
To illustrate and evaluate each component in DIGMAPPER, we use CriticalMAAS's DARPA-USGS map dataset - a set of USGS geologic maps annotated by USGS for critical mineral assessments in the following sub-sections. Some features in the dataset could contain significant annotation errors, so we curate separate map subsets for evaluating each DIGMAPPER's module in this section. In practice, DIGMAPPER has processed hundreds of geologic maps from various sources (e.g., the National Geologic Map Database (NGMDB)) to support efficient USGS-led critical mineral assessments, such as nationwide assessments of magmatic nickel-copper, lacustrine lithium, and rare earth elements (REEs), as well as regional-scale evaluations for zinc and porphyry copper. 

The DARPA-USGS map dataset comprises 48 annotated maps (i.e., with ground truth) for feature extraction (including polygon, line, and point features) and 63 maps for georeferencing. To evaluate DIGMAPPER's performance, we categorize input maps into three groups — \textit{excellent}, \textit{good}, and \textit{fair} — based on their suitability for automatic digitization. These categories help users anticipate the quality of automatic digitization results based on the characteristics of the input map. This categorization also highlights DIGMAPPER’s capability across varied map quality and feature complexities.

\subsection{Map Layout Analysis} \label{sec:layout}
\subsubsection{Method Overview}
Our map layout analysis module first segments the map into content and legend areas using a fine-tuned LayoutLMv3\footnote{\url{https://github.com/DARPA-CRITICALMAAS/uncharted-ta1/blob/main/pipelines/segmentation/README.md}}. Then the module extracts pairs of legend images and descriptions from the legend region using a prompt-based in-context learning approach (ICL) with GPT-4o~\cite{gpt4o}. This module guides the feature extraction modules by identifying which features to extract from the segmented map content area and assigning semantic labels based on the legend descriptions. 

\begin{figure}[h]
    \centering
    \includegraphics[width=\linewidth]{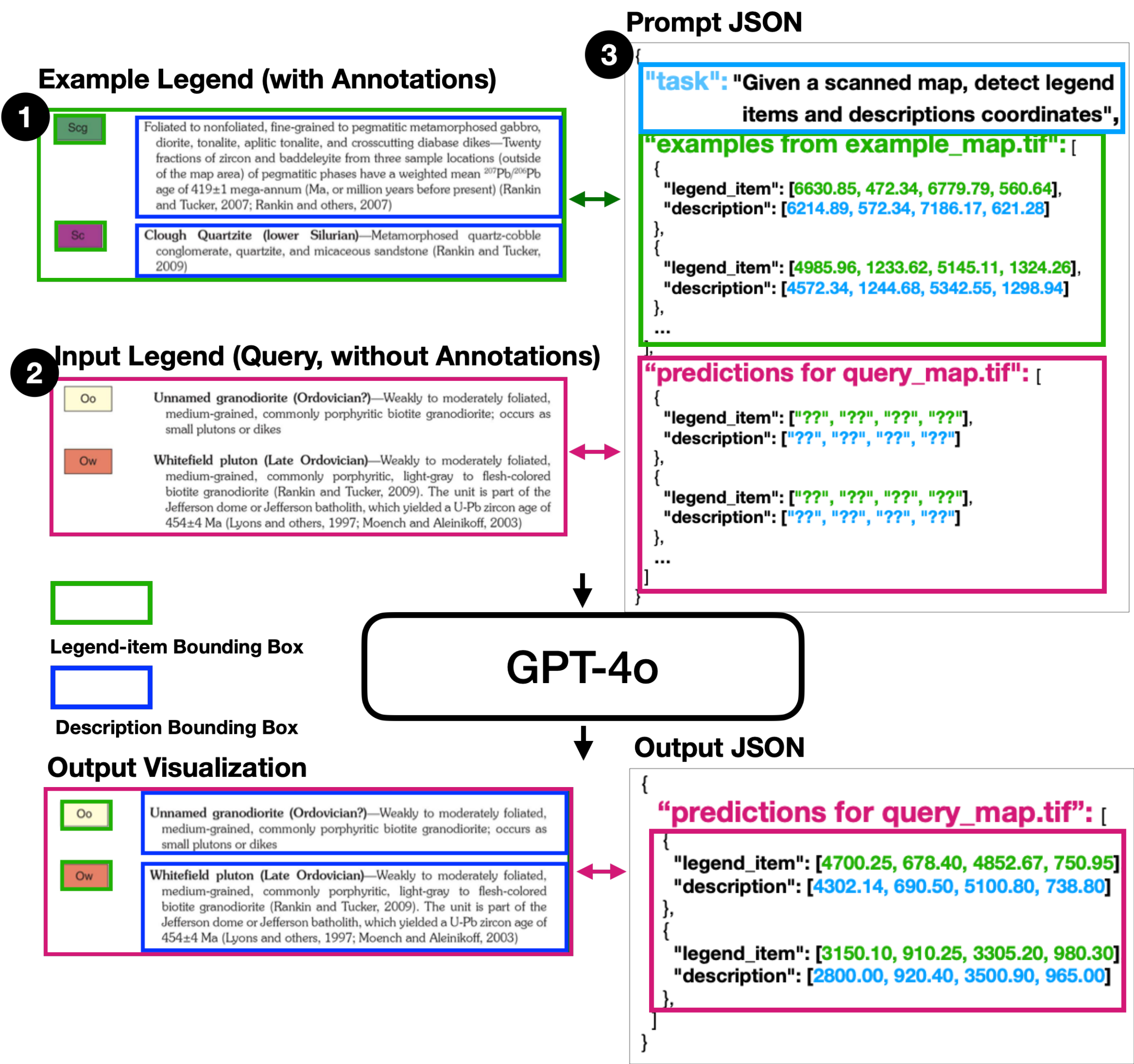}
    \caption{Workflow of our ICL approach for extracting legend-item and description pairs using GPT-4o. There are three inputs: (1) an annotated example legend area, (2) an input legend for query, and (3) a JSON-formatted prompt. The prompt includes the task definition (blue), labeled examples from the example image (green), and a query with placeholder coordinates (red). GPT-4o infers the bounding boxes for the query image. The bottom part of the figure shows the predicted legend-item and description pairs.}
    \label{fig:legend_description_workflow}
\end{figure}

For extracting pairs of legend items and corresponding descriptions, ICL enables GPT-4o to process new inputs by conditioning on task instructions and a few labeled examples provided within a prompt. Figure~\ref{fig:legend_description_workflow} shows the workflow of our ICL approach. GPT-4o takes three inputs: an example legend area (\textcircled{\small 1} in Figure~\ref{fig:legend_description_workflow}), a cropped legend area from a query map (\textcircled{\small 2} in Figure~\ref{fig:legend_description_workflow}), and a JSON-formatted prompt (\textcircled{\small 3} in Figure~\ref{fig:legend_description_workflow}). The JSON prompt includes the task instruction (highlighted as the blue box in Figure~\ref{fig:legend_description_workflow}), several annotated example pairs of legend-item and description bounding boxes from the example image (highlighted as the green box), and an unlabeled query entry with placeholder coordinates (``??'') (highlighted in the red box). GPT-4o leverages the provided examples and task definition to infer the missing bounding boxes in the query image, generating legend-item and description pairs, showing as the output in Figure~\ref{fig:legend_description_workflow}. The proposed few-shot ICL approach enables training-free adaptation to diverse map layouts.

\subsubsection{Evaluation}
For evaluation, we randomly select and manually annotate one-third of the DARPA-USGS map dataset (40 out of 111 maps), given the time-intensive nature of the annotation process. We evaluate the accuracy of our layout analysis module using two standard object detection metrics~\cite{yolov8}: Intersection over Union (IoU) and F1 score. We set an IoU threshold of 0.5 to determine true positives when calculating F1 scores. We conduct independent evaluations for map content segmentation and legend extraction.

\noindent\textbf{Map content and legend segmentation:} The second column (“Map Segment”) in Table~\ref{tab:map_layout_analysis} reports segmentation results. In the \textit{excellent} and \textit{good} categories, where maps follow conventional layouts with map content on the left and rectangular legend blocks on the right, the model achieves average IoU and F1 scores of 0.83 and 0.86, respectively. The slightly lower performance in the good category, compared to the excellent category, is largely due to minor misalignments between the predicted boxes and the ground truth, which occur when the legend areas are not strictly rectangular. The performance declines for the maps in the \textit{fair} category, which includes maps with scan noise or unconventional layouts, such as legend areas with irregular (non-rectangular) shapes. In these cases, the model often fails to segment complete regions. We expect that additional annotations and fine-tuning on maps in the fair category will improve performance.

\noindent\textbf{Legend item and description pair extraction:} We find through experiments that using 15 examples of legend-item and description pairs yields the most accurate results. The third and fourth columns of Table~\ref{tab:map_layout_analysis} present the evaluation extractions for legend items (images) and descriptions, respectively. In the \textit{excellent} and \textit{good} categories, where legend items and descriptions follow a left-right structure, and the number of legend entries is moderate, our method achieves average IoU and F1 scores of 0.85 and 0.88 for legend item extraction, and 0.85 and 0.92 for description extraction. The slightly lower performance for the maps in the good category, compared to the excellent category, is due to variations in symbol styles. The model effectively extracts colored polygon legend items (e.g., filled rectangles) but struggles with irregularly shaped black-and-white symbols. In the \textit{fair} categories, performance declines due to several challenges: extracting incomplete descriptions when they are exceptionally long; missing pairs in legends with a large number of entries (e.g., over 100); and difficulty capturing pairs in layouts where legend items and descriptions are arranged top-to-bottom rather than in the typical left-to-right structure. We expect that adding additional examples of the challenging layouts will improve the performance. In summary, our ICL method demonstrates a strong capability to extract pairs of legend items and descriptions with a typical left-right layout and promising adaptability to diverse formats without requiring extensive manual annotations.

\begin{table}[htbp]
\centering
\caption{Performance of map layout analysis module}
\label{tab:map_layout_analysis}
\begin{tabular}{l|cc|cc|cc}
\toprule
\textbf{Category} & \multicolumn{2}{c|}{\textbf{Map Segment}} & \multicolumn{2}{c|}{\textbf{Legend Item}} & \multicolumn{2}{c}{\textbf{Description}} \\
                  & \textbf{IoU} & \textbf{F1} & \textbf{IoU} & \textbf{F1} & \textbf{IoU} & \textbf{F1} \\
\midrule
Excellent         & 0.84 & 0.87 & 0.87 & 0.90 & 0.86 & 0.93 \\
Good              & 0.82 & 0.85 & 0.82 & 0.86 & 0.84 & 0.90 \\
Fair              & 0.80 & 0.84 & 0.75 & 0.78 & 0.73 & 0.77 \\
\bottomrule
\end{tabular}
\end{table}

\vspace{-10pt}
\subsection{Polygon Extraction}\label{sec:polygon}
\subsubsection{Method Overview}
\begin{figure}[t]
    \centering
    \includegraphics[width=\linewidth]{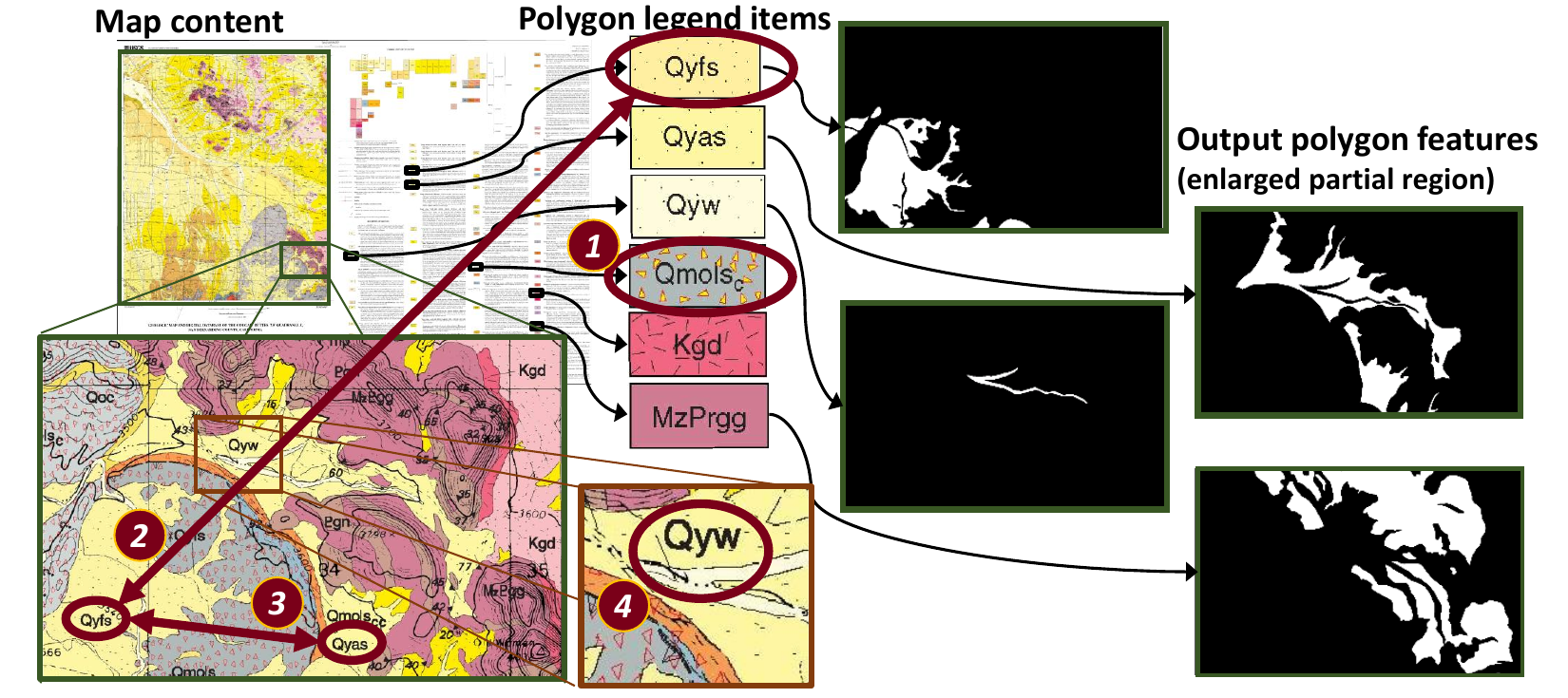}
    \caption{An example of extracting polygon features from geologic maps. There are several challenges: 1) the legend item for \emph{Qmols$_{c}$} has triangular markings; 2) \emph{Qyfs} has a color shift in its corresponding polygon features; 3) \emph{Qyfs} and \emph{Qyas} share a similar color in the map content; 4) the text label of \emph{Qyw} is located outside its polygon feature.} 
\label{fig:poly_example}
\end{figure}

Polygon features in geologic maps typically represent geologic units, such as rock formations or resource prospects. Each polygon is linked to a legend item that specifies its visual characteristics, including colors, markings, and text labels. Automatically extracting these polygon features has four challenges~\cite{lin2023exploiting} (Figure~\ref{fig:poly_example}). First, legend items come in various styles. Second, colors used in the legend items may differ from those in the corresponding map content due to scanning artifacts or overlaps with translucent symbols. Third, multiple legend items may share the same color and can only be distinguished by text labels. Fourth, small polygon size often results in text labels being placed nearby and referenced with lines rather than being embedded directly inside the feature. 

\begin{figure}[ht]
    \centering
    \includegraphics[width=\linewidth]{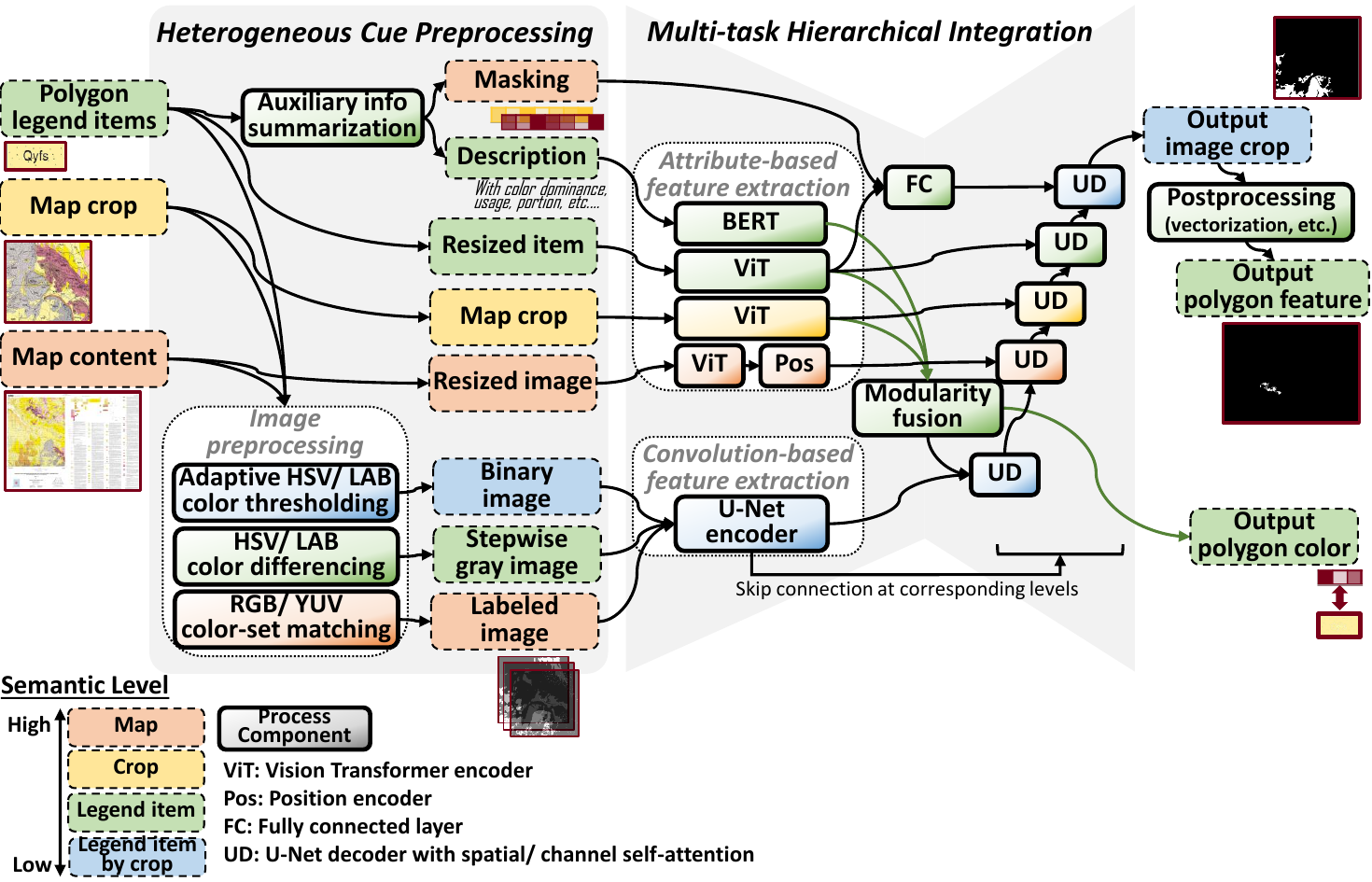}
    \caption{TOPAZ workflow for polygon feature extraction. TOPAZ outputs the vectorized polygon and its corresponding color used in the map content.} 
\label{fig:poly_workflow}
\end{figure}

To address the challenges, we propose \textbf{T}ransformer-\textbf{O}riented \textbf{P}olygon extraction with \textbf{A}daptive-hierarchical \textbf{Z}oning (TOPAZ), a multi-task U-shaped encoder-decoder model that jointly extracts polygon features, along with their exact colors in the map content, and associates them with the corresponding legend items. Figure~\ref{fig:poly_workflow} shows TOPAZ workflow, where we treat map content areas and legend items extracted by the map layout analysis module (Section~\ref{sec:layout}) as our input to generate the vectorized outputs. TOPAZ leverages heterogeneous features identified from both the map content and legend items, including color spectra (RGB, HSV, LAB, and YUV), markings, and textual descriptions. In the encoding phase, TOPAZ employs specialized encoders to extract both attribute-based and convolution-based features about polygons. Vision Transformers (ViTs), along with BERT, process the color and spatial patterns with text descriptions, while the U-Net captures the polygon topology and boundary details. Next, the decoding phase hierarchically integrates these features across distinct semantic levels, from global map-wide context to localized item-specific cues, enabling adaptive polygon refinement via a series of channel-attention mechanisms. In addition, TOPAZ enforces a one-to-one pixel-to-legend-item assignment via polygon masking, and applies multi-task training to simultaneously optimize polygon shape extraction and color recognition with the corresponding legend item. This design allows TOPAZ to generate vectorized polygon features explicitly paired with intended legend items under degraded visual conditions.

\subsubsection{Evaluation}
For evaluation, we select 11 maps containing 221 polygon legend items from the DARPA-USGS map dataset by excluding two maps with significant annotation errors from 13 maps with annotated polygon ground truth. We use both pixel-level and instance-level metrics, corresponding to evaluations in raster and vector formats, respectively. Pixel-level metrics (raster performance) focus on boundary alignment and fine-grained pixel accuracy in raster space, while instance-level metrics (vector performance) evaluate the structural consistency and topological correctness of polygons in vector space, which are critical for map digitization and can affect downstream application efficiency. 

Pixel-level metrics include Intersection over Union (IoU) and F1 score, both of which range from 0 to 1, where higher values indicate better raster performance. IoU measures the pixel overlap between extracted polygons and ground truth, reflecting shape accuracy. F1 score is the harmonic mean of precision and recall based on pixel counts, capturing general correctness. Instance-level metrics include the candidate (cand) ratio and extraction-ground-truth (ex-gt) ratio, both of which range from 0 to infinity, where values closer to 1 indicate better vector performance. The cand ratio refers to the number of extracted polygons over ground truth, indicating local topology accuracy by penalizing redundant extractions. The ex-gt ratio refers to the total number of extracted polygons over ground truth, assessing global topology accuracy.

All reported metrics are weighted by the IoU count, emphasizing the geometric and topological alignment between the extracted and ground truth polygons. The IoU count refers to the number of intersected polygons between the extracted results and the ground truth, plus the number of intersections between ground truth polygons and non-extracted areas, both computed after vectorization. 

Table~\ref{tab:poly_evaluation} presents the evaluation results for polygon feature extraction across different map categories. For maps listed under the \textit{excellent} and \textit{good} categories, our approach demonstrates good accuracy in raster and vector metrics, achieving an F1 score of 0.98 and an Ex-gt ratio of 0.99, respectively. Maps in the \textit{excellent} and \textit{good} categories typically have limited visual interference from shaded relief or scanning artifacts, with maps for \textit{excellent} containing fewer fragmented polygons and maps for \textit{good} having more fragmented polygons, as reported in the IoU count. In both cases, regardless of the underlying map styles, color usage across legend items, or the presence of opaque features, TOPAZ can extract polygons that align well with ground truth boundaries, with instance counts closely matching the reference.

We identify two characteristics of maps that contribute to the maps categorized as \textit{fair}: fragmented polygons and significant color shifts. First, since performance is weighted by IoU count, missing even small polygons in fragmented ground truth can disproportionately decrease overall performance. This is supported by the fact that removing IoU-count weighting can lead to an increase of at least 0.30 in both IoU and F1 score for these maps. Second, significant color shifts between legend items and corresponding map content, which are often caused by overlaps with shaded relief or elevation models, can result in fragmented extraction of a single ground-truth polygon. This fragmentation increases the IoU count and decreases the weighted performance.

\begin{table}[htbp]
\small
\centering
\caption{Evaluation results for polygon features. The table reports raster and vector metrics across different map categories. All the performance is weighted by IoU count.}
\label{tab:poly_evaluation}
\begin{tabular}{c|c c c c|c c}
\hline
\textbf{Category} & \multicolumn{4}{c|}{\textbf{Evaluation Metrics}} & \multicolumn{2}{c}{\textbf{Statistics}} \\ 
\cline{2-7}
 & \textbf{Median} & \textbf{Median} & \textbf{Cand} & \textbf{Ex-gt} & \textbf{Map} & \textbf{IoU} \\
 & \textbf{IoU} & \textbf{F1 Score} & \textbf{Ratio} & \textbf{Ratio} & \textbf{Count} & \textbf{Count} \\
\hline
Excellent & \textbf{0.95} & \textbf{0.98} & \textbf{8.54} & 1.31 & 2 & 14.16 \\
Good & 0.63 & 0.77 & 14.60 & \textbf{0.99} & 5 & 23.99 \\
Fair & 0.17 & 0.28 & 41.63 & 1.86 & 4 & 36.78 \\ \hline
\end{tabular}
\end{table}

\subsection{Line Extraction}\label{sec:line}
\subsubsection{Method Overview}
Automatically extracting linear features, such as fault lines, from geologic maps encounters two key challenges: capturing adequate image context and spatial context. Insufficient image context leads to false detections by failing to distinguish desired linear objects from others with similar appearances. Meanwhile, insufficient spatial context hampers the accurate delineation of elongated, slender-shaped linear objects. 

\begin{figure}[ht]
    \centering
    \includegraphics[width=\linewidth]{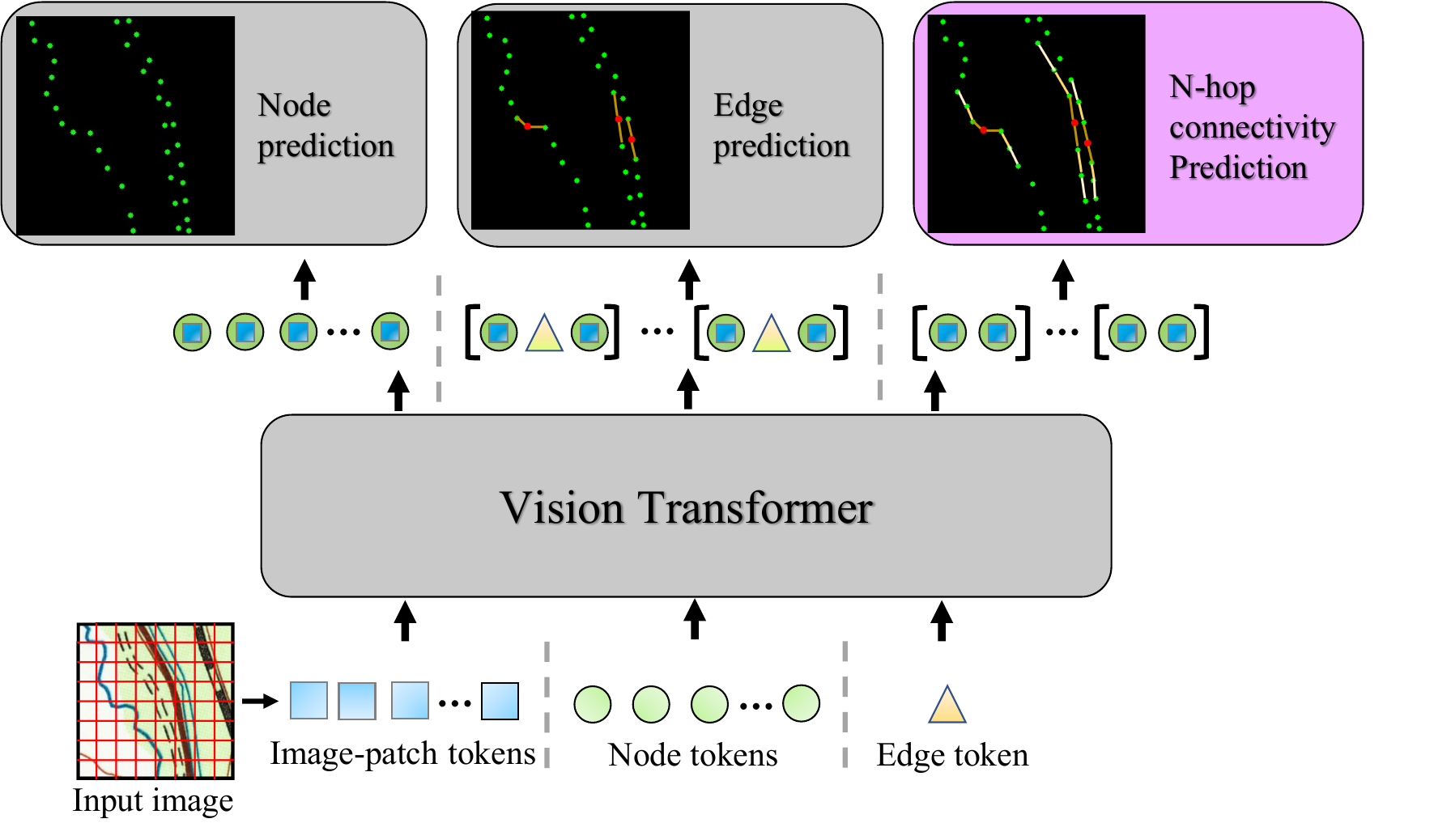}
    \caption{LDTR Overview. LDTR inputs images and outputs the graph for the desired lines by predicting nodes and edges. Notably, the N-hop connectivity prediction head in LDTR is crucial for capturing the complex spatial context among nodes and thereby enhancing the connectivity of the detected lines. Best viewed in color.} 
\label{fig:ldtr_overview}
\end{figure}

To address both challenges, we propose the \textbf{L}inear Object \textbf{D}etection \textbf{TR}ansformer (LDTR), which extends Relationformer~\cite{relationformer} with a novel N-hop connectivity prediction module. The module enables nodes to aggregate spatial context from both adjacent and distant nodes, enhancing the model’s ability to capture line orientation, curvature, and overall topology. Additionally, LDTR leverages deformable attention~\cite{deform-detr} to selectively focus on representative patterns along the desired lines, helping distinguish desired lines from visually similar, non-desired lines. Figure~\ref{fig:ldtr_overview} illustrates the architecture overview of LDTR, which processes image-patch, node, and edge tokens to construct accurate vector graphs by integrating representative visual patterns with rich spatial context.

\subsubsection{Evaluation}
\begin{figure}[htbp]
\centering
\small 
\begin{tabular}{@{}c@{\hspace{1pt}}c@{\hspace{1pt}}c@{}} 
\fbox{\includegraphics[width=0.30\linewidth]{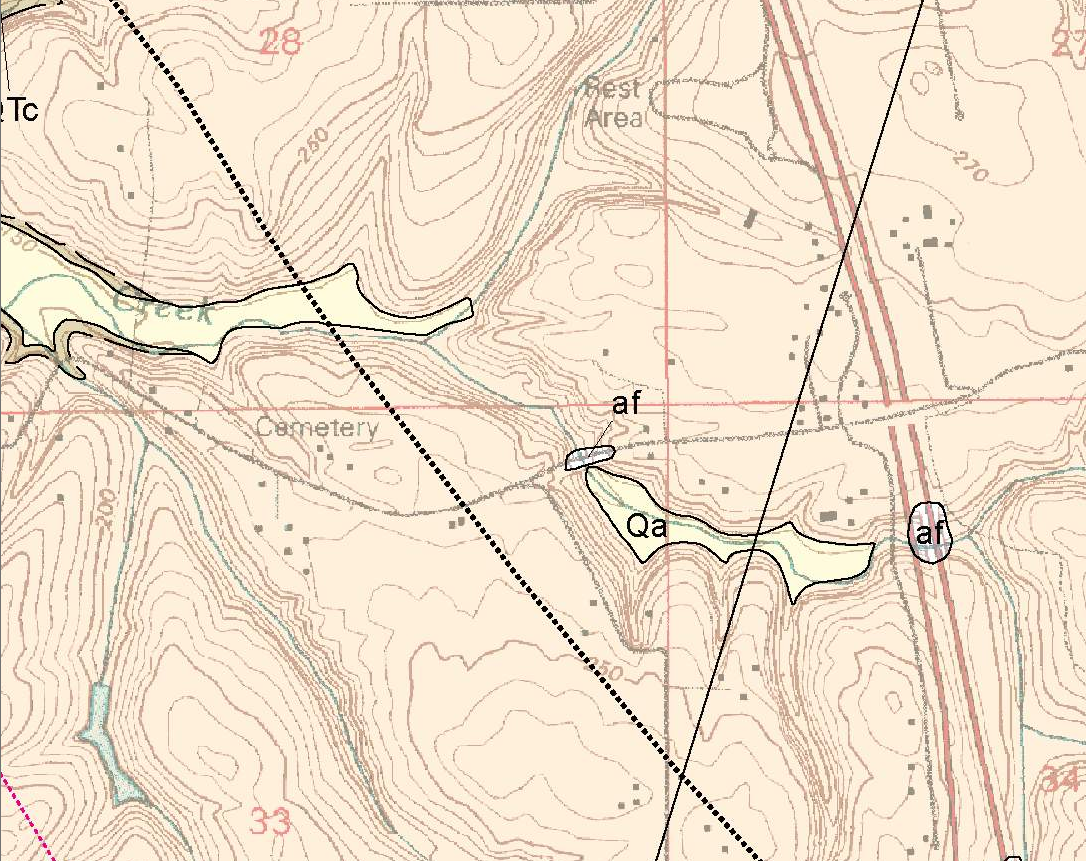}}&
\fbox{\includegraphics[width=0.30\linewidth]{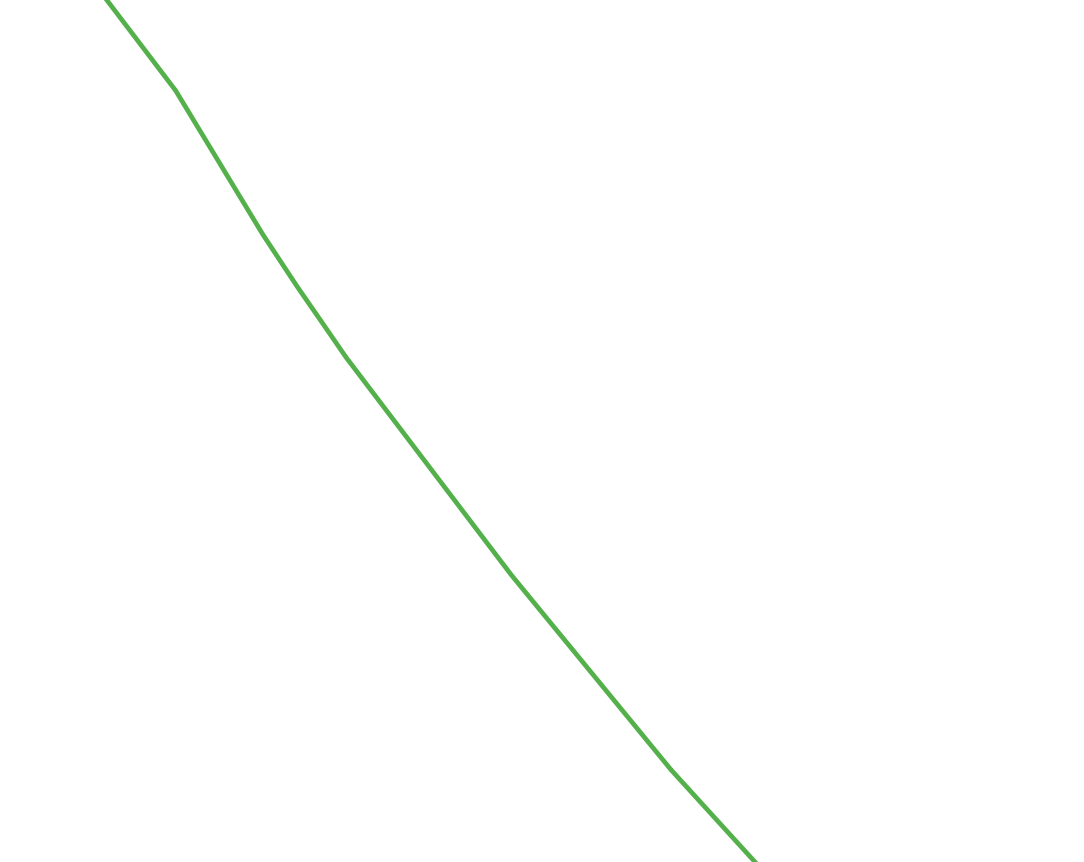}} &
\fbox{\includegraphics[width=0.30\linewidth]{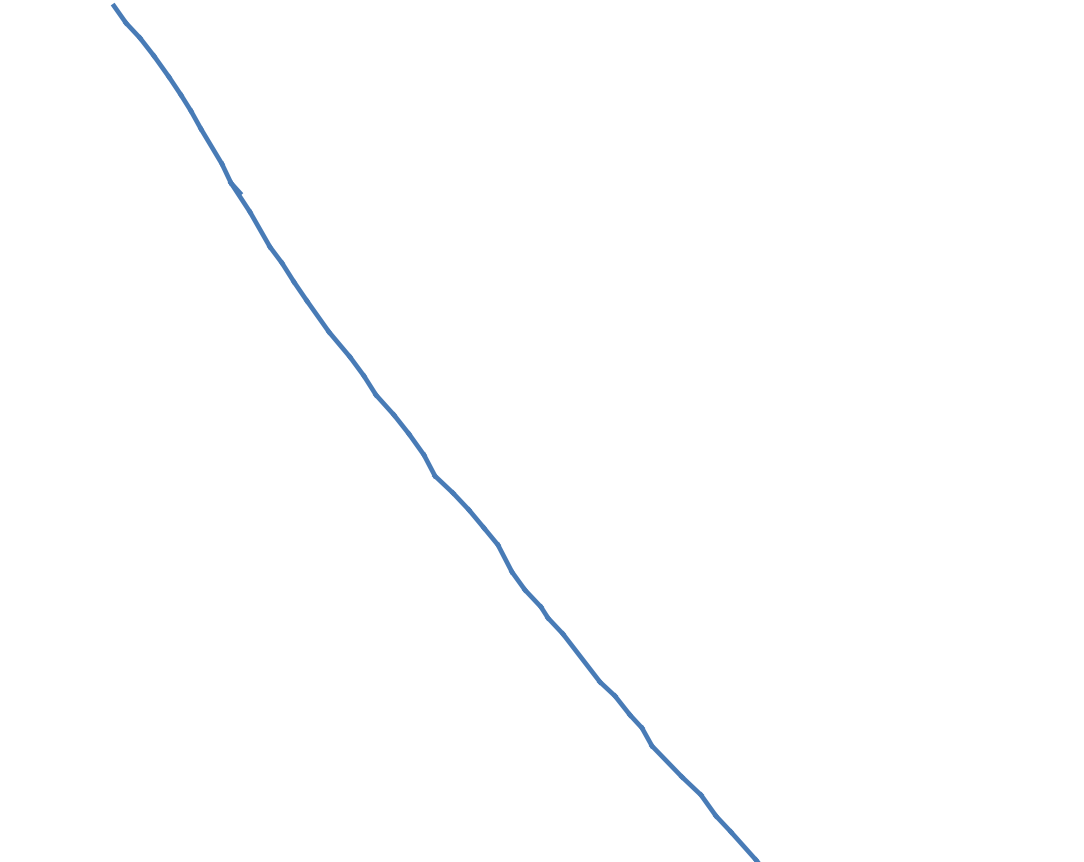}} \\
\fbox{\includegraphics[width=0.30\linewidth]{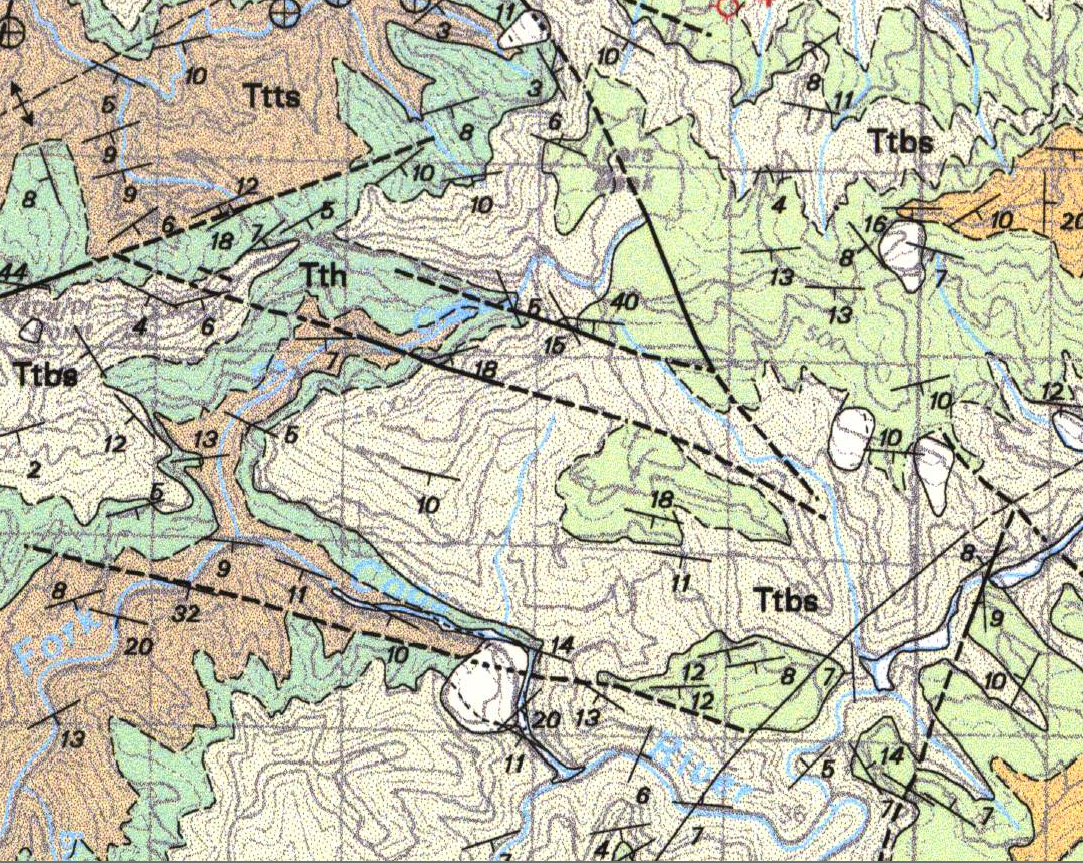}} &
\fbox{\includegraphics[width=0.30\linewidth]{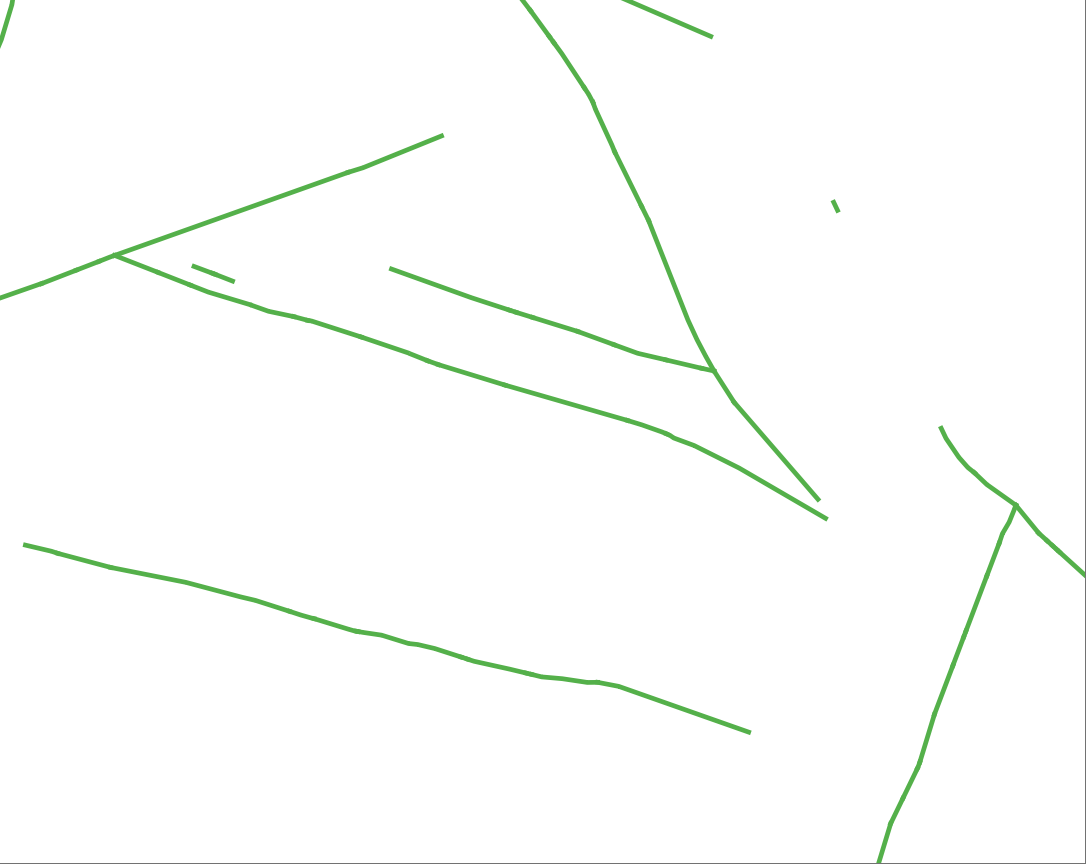}} &
\fbox{\includegraphics[width=0.30\linewidth]{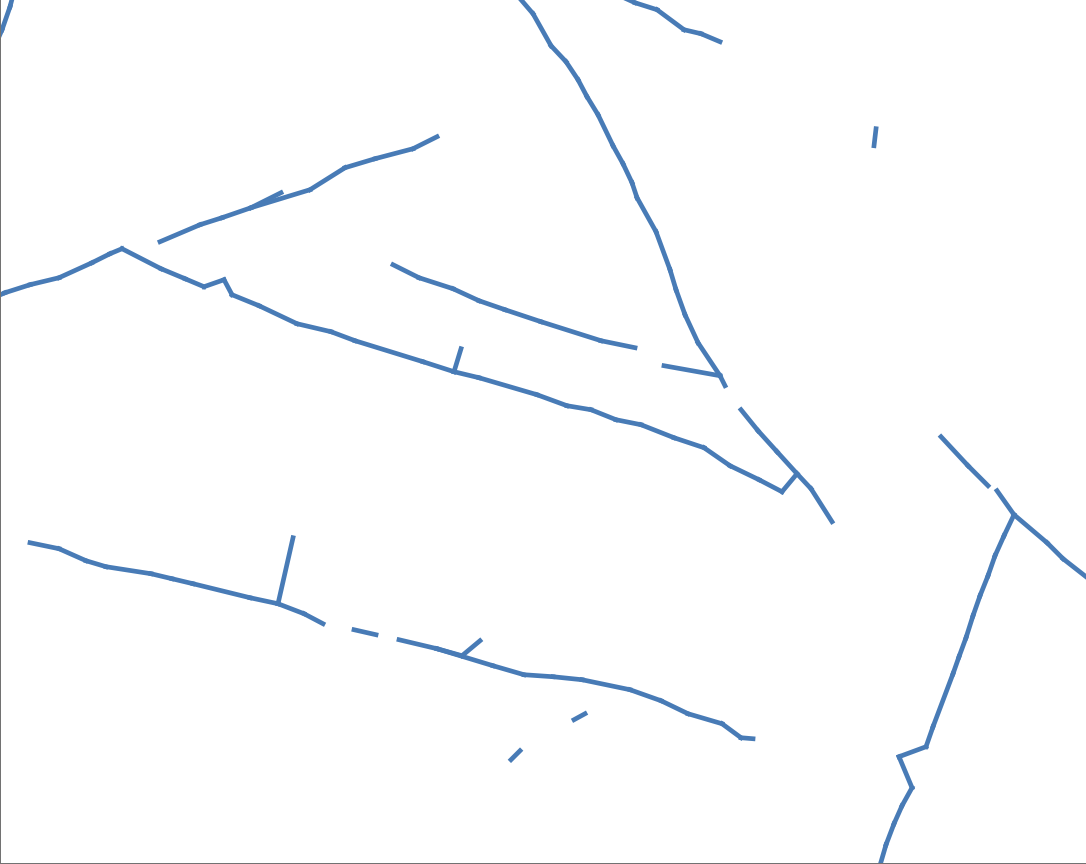}} \\
\fbox{\includegraphics[width=0.30\linewidth]{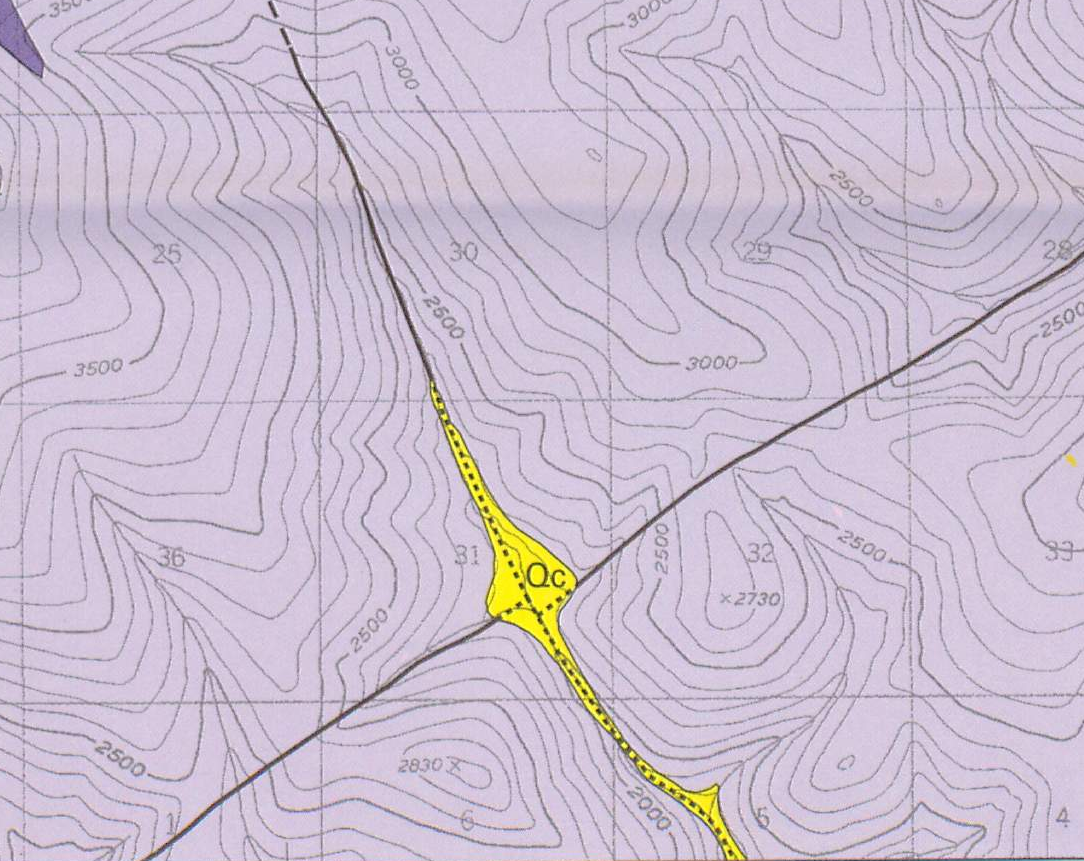}} &
\fbox{\includegraphics[width=0.30\linewidth]{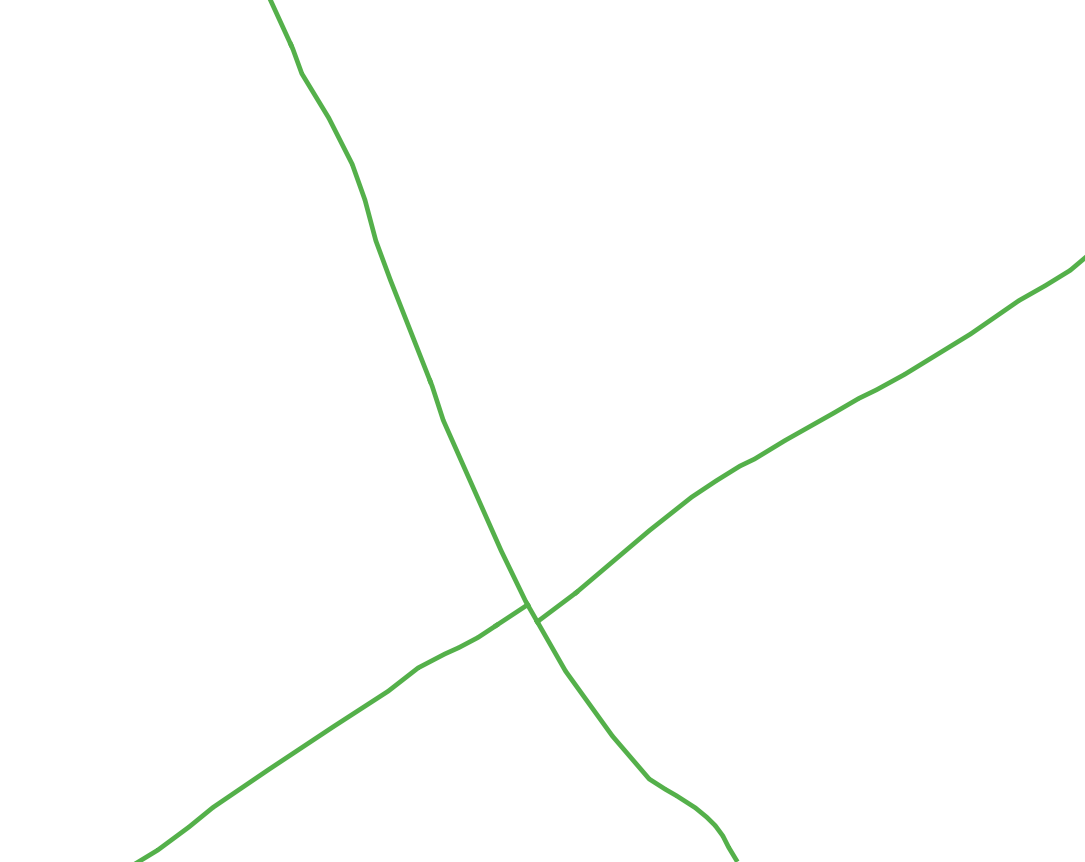}} &
\fbox{\includegraphics[width=0.30\linewidth]{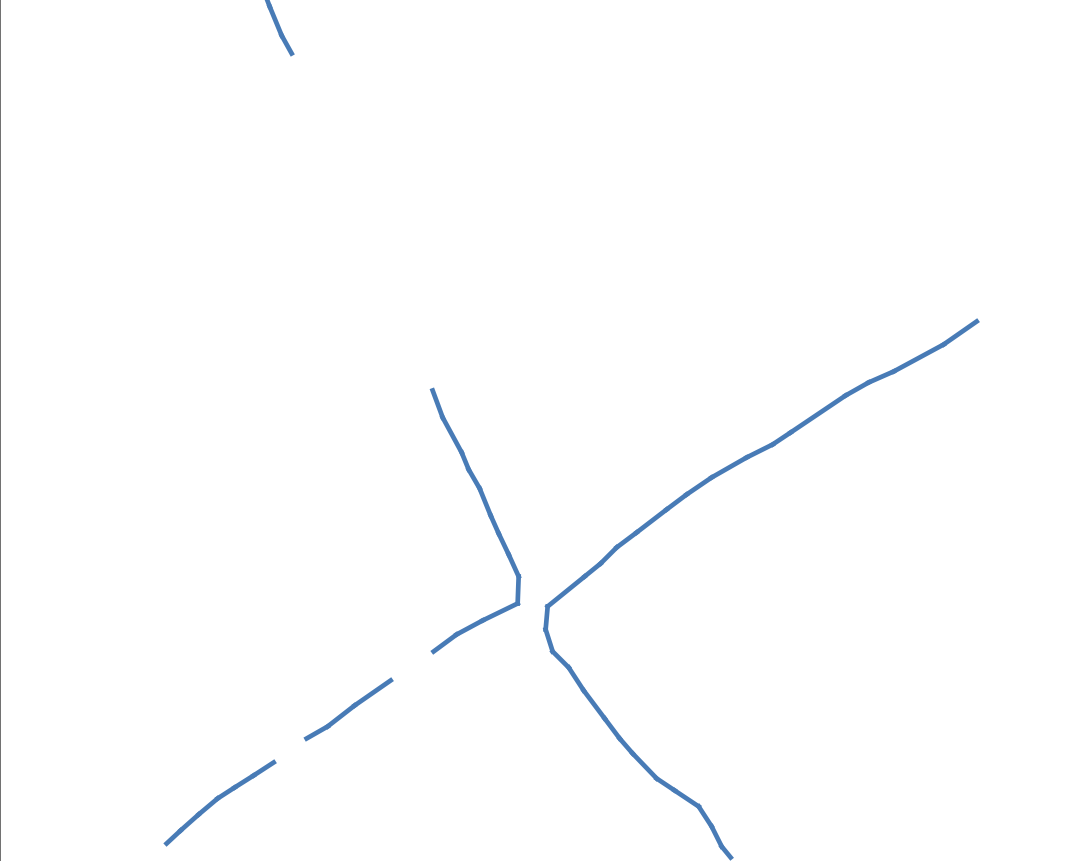}} \\
\multicolumn{1}{c}{\textbf{Map}} &
\multicolumn{1}{c}{\textbf{Ground Truth}} &
\multicolumn{1}{c}{\textbf{Extraction}} \\
\end{tabular}
\caption{Examples of fault line extraction results across performance categories. Each row corresponds to one map from the excellent, good, and fair groups, respectively, and shows (from left to right) the map image, ground truth, and the extracted fault lines from LDTR.}
\label{fig:fault_extraction_examples}
\end{figure}

For evaluation, we focus on two linear features: fault lines and thrust fault lines. Both types of faults are essential to mineral assessment, as they often serve as pathways for hydrothermal fluids responsible for concentrating and depositing valuable mineral resources~\cite{fault_importance}. Fault lines appear as thick black lines in solid, dotted, or dashed patterns, showing in the first column of Figure~\ref{fig:fault_extraction_examples}. Thrust fault lines follow the style but are distinguished by black triangles along the line. The evaluation dataset consists of 10 maps selected from the DARPA-USGS map dataset, excluding those with severe annotation errors. To assess performance, we used correctness and completeness~\cite{correct_complete}, which respectively measures how accurately and completely the extracted lines align with the ground truth. 

\begin{figure}[htbp]
\centering
\setlength{\belowcaptionskip}{-1pt} 
\begin{subfigure}[t]{0.45\linewidth}
    \centering
    \fbox{\includegraphics[width=\linewidth]{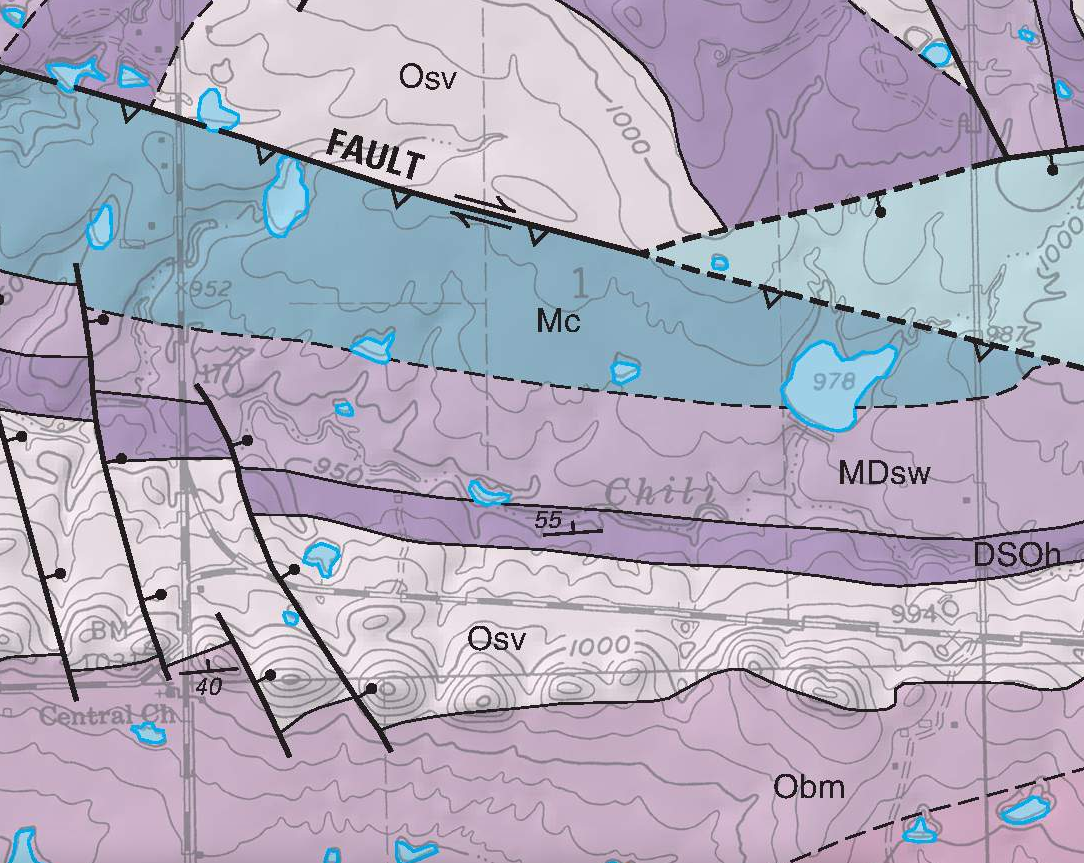}}
    \caption{Map}
    \label{fig:line_refinement_map}
\end{subfigure}
\hfill
\begin{subfigure}[t]{0.45\linewidth}
    \centering
    \fbox{\includegraphics[width=\linewidth]{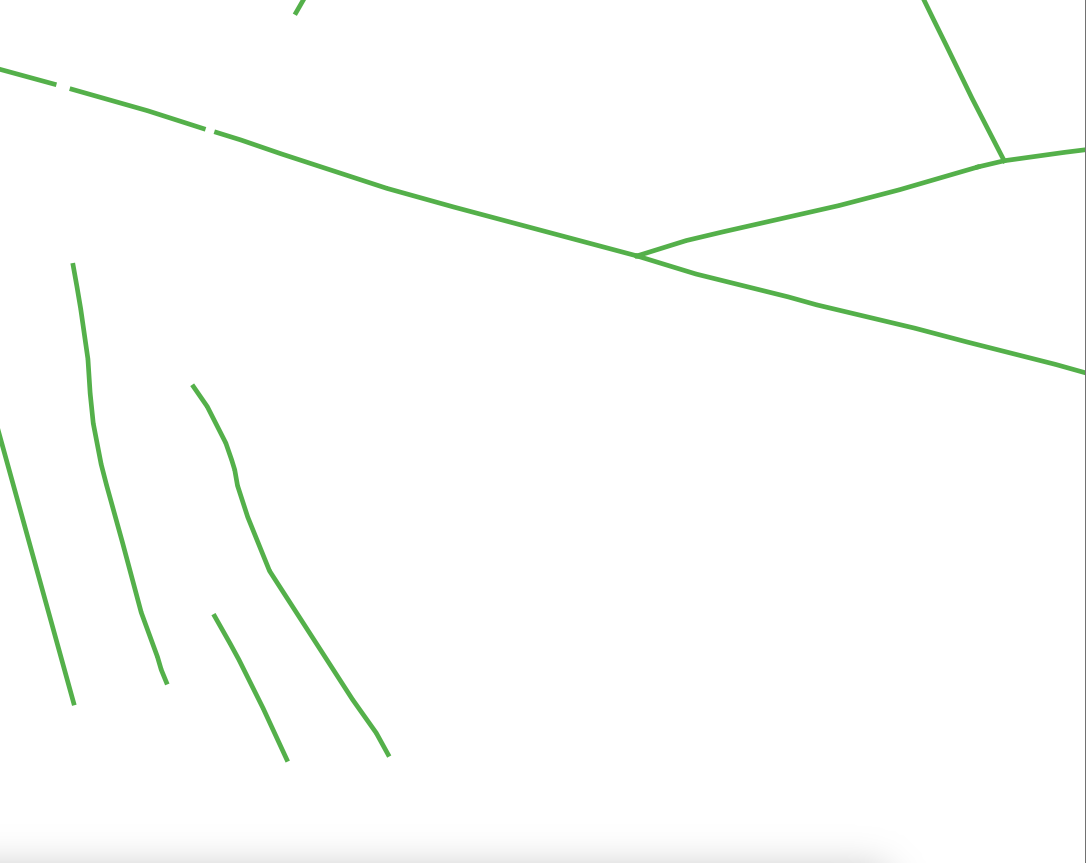}}
    \caption{Ground Truth}
    \label{fig:line_refinement_gt}
\end{subfigure}

\begin{subfigure}[t]{0.45\linewidth}
    \centering
    \fbox{\includegraphics[width=\linewidth]{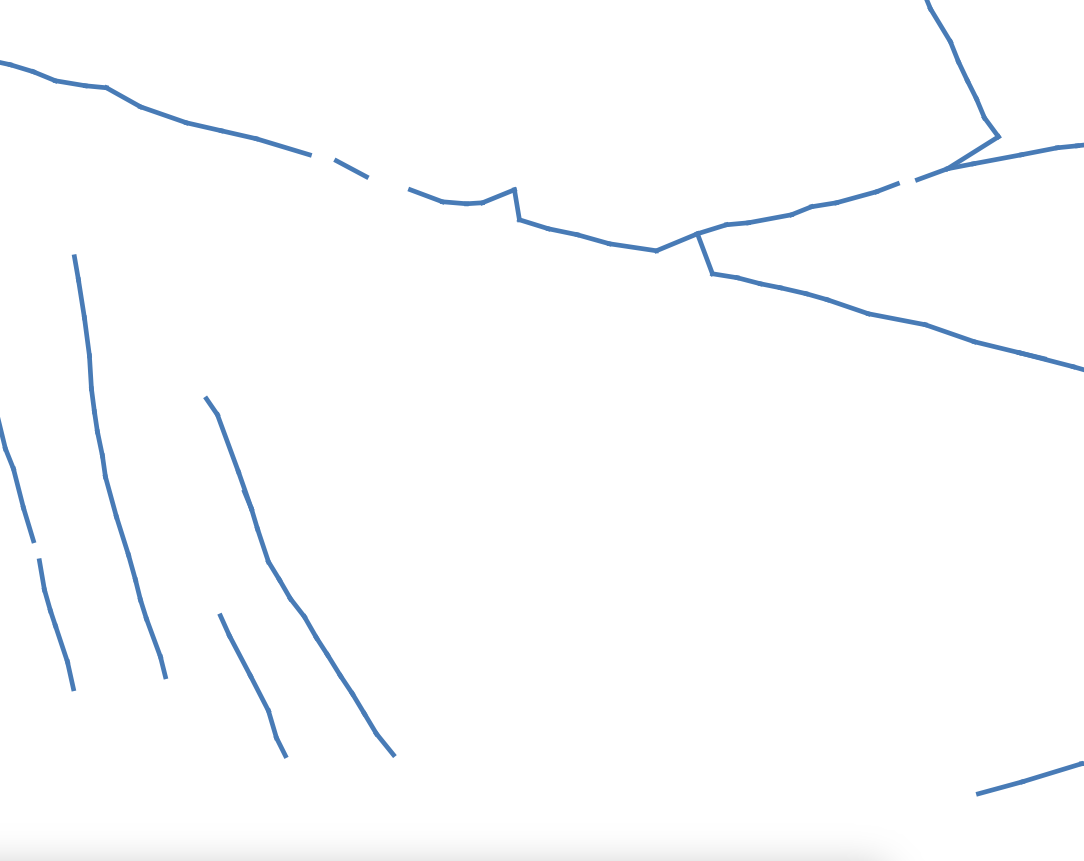}}
    \caption{Extracted Lines from LDTR}
    \label{fig:line_refinement_extraction}
\end{subfigure}
\hfill
\begin{subfigure}[t]{0.45\linewidth}
    \centering
    \fbox{\includegraphics[width=\linewidth]{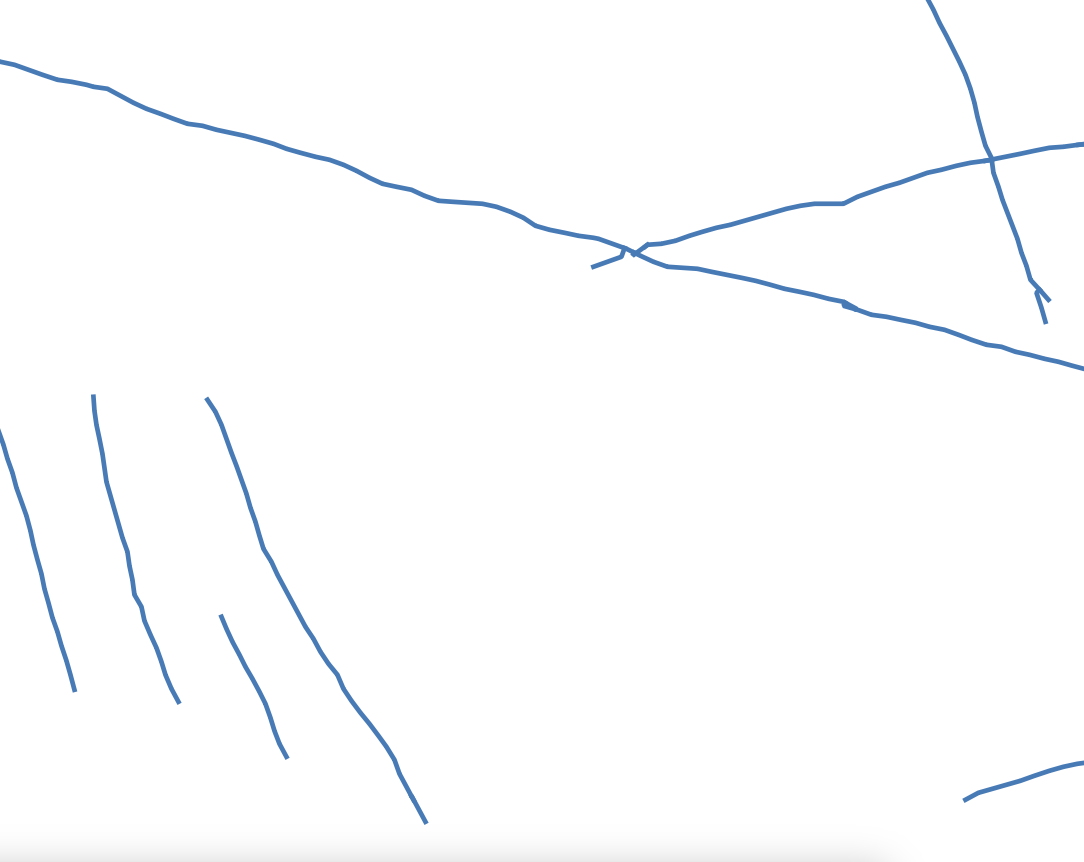}}
    \caption{Refined Lines from (c)}
    \label{fig:line_refinement_refinement}
\end{subfigure}
\caption{Example of line refinement using a language model. (c) shows initial extracted lines using LDTR. (d) present the refined lines from (c), produced by a language-based model that enhances line geometric and topological accuracy.}
\label{fig:line_refinement}
\end{figure}

Table~\ref{tab:line_evaluation} presents the evaluation results for fault lines and thrust fault lines across three performance categories. On high-quality digital-born or well-scanned maps in the \textit{excellent} group, LDTR achieves high correctness and completeness (up to 0.88 and 0.95 for fault lines; 0.94 and 0.99 for thrust faults). In the \textit{excellent} group, the maps contain distinguishable features, and the extracted lines are ready for downstream use, showing in the first row of Figure~\ref{fig:fault_extraction_examples}. Maps in the \textit{good} category, which exhibit mild scan inconsistencies, still yield 0.73 correctness and 0.93 completeness for fault lines. Visual artifacts such as color granularity and background inconsistency (second row in Figure~\ref{fig:fault_extraction_examples}) lead to incorrect line connections. The extraction in the \textit{good} category requires human curation to ensure optimal quality for downstream use. In the \textit{fair} category, performance further declines, primarily due to faded or ambiguous symbols and poor map scan quality, leading to missed detections (third row in Figure~\ref{fig:fault_extraction_examples}). We expect to add annotated data from low-quality scanned maps to improve extraction accuracy. To further reduce manual work on refining extracted line geometry and topology, we plan to leverage the architecture of language models~\cite{lewis2019bart} to enhance accuracy similar to trajectory refinement~\cite{trajectory_refinement}. By capturing sequential patterns, language models can improve understanding of line topologies and geometric characteristics~\cite{trajectory_refinement}. Figure~\ref{fig:line_refinement_refinement} shows the preliminary results of line refinement using Figure~\ref{fig:line_refinement_extraction} as input. The refined lines show enhanced geometric and topological accuracy, notably reducing discontinuities, smoothing zigzag artifacts, and improving line intersections. In summary, LDTR demonstrates strong performance in extracting fault and thrust fault lines on high-quality maps, with ongoing efforts focused on improving accuracy and reducing manual refinement for maps with mild scan noise.

\begin{table}[htbp]
\small
\centering
\caption{Evaluation results for fault and thrust fault lines.}
\label{tab:line_evaluation}
\begin{tabular}{l|c|c}
\hline
\textbf{Feature \& Category} & \textbf{Correctness} & \textbf{Completeness} \\
\hline
\textbf{Fault line (Excellent)} & \textbf{0.88} & \textbf{0.95} \\
Fault line (Good) & 0.73 & 0.93 \\
Fault line (Fair) & 0.40 & 0.24 \\
\hline
\textbf{Thrust fault (Excellent)} & \textbf{0.94} & \textbf{0.99} \\
Thrust fault (Good) & 0.98 & 0.42 \\
Thrust fault (Fair) & 0.02 & 0.01 \\
\hline
\end{tabular}
\end{table}

\setlength{\fboxsep}{1pt} 

\subsection{Point Extraction}\label{sec:point}
\subsubsection{Method Overview} 
\begin{figure}[ht]
    \centering
    \includegraphics[width=0.8\linewidth]{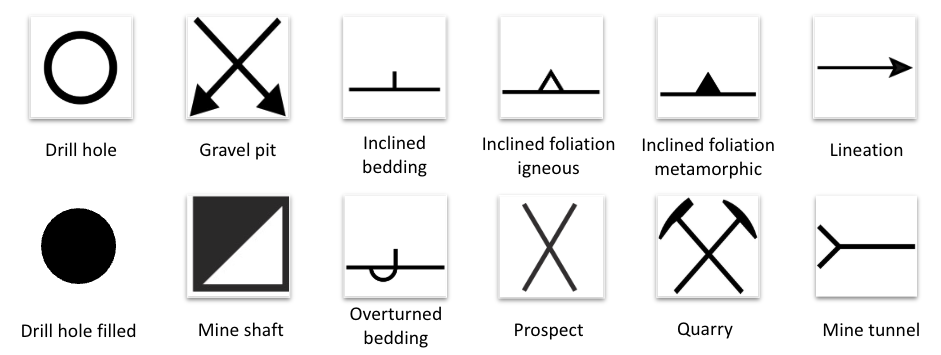}
    \caption{Twelve predefined point symbols} 
\label{fig:point_catalog}
\end{figure}

Extracting point features (e.g., mining site symbols) using machine learning models is challenging due to the limited availability of annotations for training, as well as the high variability in the size, orientation, and shape of point symbols. We developed an object detection approach based on YOLO-v8~\cite{yolov8}, which incorporates synthetic and human annotations for training to mitigate the reliance on extensive human annotations. The model aims to detect 12 predefined point symbols (Figure~\ref{fig:point_catalog}) selected based on actual usage by USGS geologists. 

To address limited training data, we propose a synthetic data generation approach consisting of three steps: (1) Given the predefined set of point symbols, our approach uses a GPT-based method to automatically identify table rows in \cite{subcommittee2006fgdc} that contain both symbol images and their textual descriptions, and then extracts the corresponding symbol illustrations from those rows; (2) Our method generates base maps by segmenting map content areas using the layout analysis module (Section~\ref{sec:layout}) and masking out annotated symbol regions; (3) Our method simulate realistic map compositions by randomly overlaying the extracted symbols onto the base maps. To enhance the variability of synthetic images, we apply random rotations, rescaling, and blurring to the symbol images. Additionally, for symbols that typically include directional indicators with numbers (e.g., bedding symbols), we also place randomly generated numbers nearby to mimic their typical context. Given the large map size (over 10,000 × 10,000 pixels), our point module crops maps into smaller 1,000 × 1,000-pixel patches for YOLOv8 training. In total, we generate 10,000 synthetic patches, with approximately 3,000 patches per symbol class for training. In addition to synthetic data, we incorporate human annotations from 100 USGS geologic maps, and train YOLO-v8 on the combined dataset of synthetic and human-labeled data for point symbol detection. 

\subsubsection{Evaluation} 
We evaluate model performance using 10 maps containing five symbol types from the DARPA-USGS map dataset. We exclude the maps that do not have human-annotated point symbols or contain incorrect annotations to ensure a reliable evaluation. The other seven symbol types, although they exist in the training data, do not have reliable ground truth in the DARPA-USGS map dataset. We use instance-level precision, recall, and F1-score for evaluation. A detected symbol is considered correct if it falls within a spatial buffer around the ground truth point, defined as $2\times10^{-4} \times$ map-diagonal-length (in pixels), which is approximately the point symbol size for each map. 

Table \ref{tab:point_eval_category} presents the map-level performance. We categorize the 10 maps into three performance levels: \textit{Excellent}, \textit{Good}, or \textit{Fair}, based on their F1 scores. We observe that maps in the \textit{Excellent} category typically exhibit high visual clarity, including digitally born and high-resolution scanned maps. The maps in the \textit{Good} category often have complex or dark backgrounds, making it difficult to distinguish point symbols. The maps in the \textit{Fair} category contain mostly blurred, distorted, and underrepresented point symbols.

\begin{table}[htbp]
\scriptsize \setlength{\tabcolsep}{3pt}
\small
\centering
\caption{Precision, recall, and F1 score for maps with three performance categories: Excellent, Good, and Fair}
\label{tab:point_eval_category}
\begin{tabular}{p{1.2cm} >{\centering\arraybackslash}p{1.2cm} >{\centering\arraybackslash}p{1cm} >{\centering\arraybackslash}p{1cm} >{\centering\arraybackslash}p{1cm}}
\hline
\textbf{Category}& \textbf{\# of maps} & \textbf{Precison} & \textbf{Recall}  & \textbf{F1} 
\\
\hline
Excellent & 3 &
0.90 & 0.88 & 0.89
\\
\hline
Good & 5 & 
0.58 & 0.55 & 0.56
\\
\hline
Fair & 2 & 
0.25 & 0.03 & 0.05
\\
\hline
\end{tabular}
\end{table}

Table~\ref{tab:point_evaluation} reports the model performance for each symbol type in the test maps, in which the overall performance is computed by treating all test instances equally across symbol types. Overall, our model achieves an F1 score of 0.82 across 10 maps. Furthermore, inclined bedding shows the highest performance while inclined foliation metamorphic ranks second with an F1 score of 0.69, likely due to a significantly smaller number of human-labeled training samples than inclined bedding (10,635 vs. 1,887). Overturned bedding achieves an F1 score of 0.41 with high recall and low precision. Most false positives arise from confusion with inclined bedding (e.g., inclined bedding intersects polygon features). Lineation and inclined foliation igneous have a low recall, largely because of the extremely limited number of human-labeled training samples. Additionally, lineation has varying shapes and colors across maps, such as solid arrows or hollow arrows with triangular tips. Inclined foliation igneous often appears distorted or blurred in maps with dark or densely textured backgrounds.

\begin{table}[htbp]
\scriptsize  
\setlength{\tabcolsep}{1.8pt}
\small
\centering
\caption{Precision (P), recall (R), and F1 score for five point symbols. ``\#Train`` and ``\#Eval`` indicate the number of human-label instances used for training and evaluation. ``Overall'' performance is measured by treating all test instances equally. The ``human-only baseline'' refers to a model trained only with human annotations.}
\begin{tabular}{p{3.7cm} | >{\centering\arraybackslash}p{1cm} |
>{\centering\arraybackslash}p{1cm} | >{\centering\arraybackslash}p{0.6cm} >{\centering\arraybackslash}p{0.6cm} >{\centering\arraybackslash}p{0.6cm}}

\hline
\textbf{Point Symbol} & \textbf{\#Train}  & \textbf{\#Eval} & \textbf{P} & \textbf{R} & \textbf{F1} \\
\hline
Inclined bedding & 10,625 & 1,665 & 0.90 & 0.91 & 0.91 \\
Inclined foliation metamorphic & 1,887 & 595 & 0.71 & 0.67 & 0.69 \\
Overturned bedding & 214 & 23 & 0.29 & 0.69 & 0.41 \\
Lineation & 57 & 75 & 0.24 & 0.09 & 0.13 \\
Inclined foliation igneous & 127 & 46  & 0.44 & 0.08 & 0.14 \\
\hline \hline
Overall & -- & 2,404 & \textbf{0.83} & \textbf{0.81} & \textbf{0.82} \\
\hline
Overall (human-only baseline) & -- &  2,404 & 0.77 & 0.80 & 0.79 \\
\hline
\end{tabular}
\label{tab:point_evaluation}
\end{table}

We also conduct an ablation study to examine the impact of synthetic data by comparing our model to a baseline trained solely with human annotations (the ``human-only baseline''). The last two rows of Table~\ref{tab:point_evaluation} show that incorporating synthetic data improves the overall F1 score by 3\%, helping to mitigate the challenge of limited human annotations. While a 3\% overall gain may appear modest—due to the test set being dominated by inclined bedding, which already achieves strong performance without synthetic data, there are substantial improvements for specific symbol types. In particular, F1 scores for overturned bedding and lineation increase by 19\% and 13\%, respectively. Notably, the baseline model trained on only 57 samples fails to detect any lineation instances. Still, the domain gaps between synthetic and real-world maps could limit the effectiveness of synthetic data. The synthetic symbol images are cropped from official documents~\cite{subcommittee2006fgdc}, while real-world maps often include handwritten or non-standard shapes that deviate from standard designs. The random placement of symbols in synthetic data also overlooks co-occurrence patterns and spatial relationships with other map features (e.g., lines), leading to stylistic differences between synthetic and real-world maps.

In summary, the point extraction module demonstrates strong performance on digital-born and high-resolution scanned maps. These maps enable our point model to reliably detect point symbols, particularly those with more training data and less ambiguity (e.g., inclined bedding). Incorporating synthetic data helps improve overall detection accuracy. Further enhancing the diversity of synthetic data or applying few-shot learning techniques can help strengthen the model's performance on underrepresented symbol types.

\subsection{Georeferencing}\label{sec:georeference}
Georeferencing is the process of determining the geographical extent of the map content. The output of this process is a set of ground control points (GCPs), each linking a pixel location on the map to a geographic location on the Earth’s surface. We develop two complementary approaches to georeferencing. The text-based method (Section ~\ref{subsec:georef-geocoord}) leverages textual elements, i.e., geocoordinate annotations and place names, from the map to directly generate GCP pairs or to find matched known locations using metadata repositories. When the text-based approach cannot confidently determine a concrete geographic location of an input geologic map, the visual-based approach (Section ~\ref{subsec:georef-viz}) uses the information extracted from the text-based method to construct a set of candidate georeferenced USGS topographic maps that may overlap with the geologic map. Then, the visual-based method captures and compares visual features between the input geologic map and each candidate topographic map to determine the best match for georeferencing.
\subsubsection{Text-based Method}
\label{subsec:georef-text}
We propose a text-based ensemble approach that integrates (1) geocoordinate extraction at the map content area corners and (2) place name recognition from the map title for topographic map retrieval. 

\noindent \textbf{Geocoordinate Extraction:} \label{subsec:georef-geocoord}
For maps with rectangular map content areas, geocoordinate labels are typically positioned at the four corners of the content areas. To locate these corners, DIGMAPPER first applies the map layout analysis module (Section~\ref{sec:layout}) to identify the minimum bounding box of the content region. While generally effective, the layout analysis may yield imprecise corner estimates, returning points near but not exactly at the true corners. To ensure the presence of geocoordinate labels, we define a buffer region of size 1,000 × 1,000 pixels centered around each detected corner. Within this region, we implement a corner point refinement algorithm that integrates intensity thresholding, morphological filtering, and Hough Transform-based line detection. The detected corner points are treated as source ground control points (GCPs). To obtain corresponding target GCPs in geographic coordinates, DIGMAPPER uses the finetuned text spotter, Palette \cite{lin2024hyper}, in mapKurator \cite{10.1145/3589132.3625579, li2020automatic} to recognize latitude and longitude labels within the cropped corner patches. DIGMAPPER incorporates a heuristic that assumes each map contains four corners and, consequently, eight geocoordinate labels — two per corner (one latitude and one longitude). Given that maps typically follow canonical orientation, we further constrain the geocoordinate labels to represent exactly four unique degree values: two distinct latitudes and two distinct longitudes. If the extracted geocoordinates satisfy these criteria, they are used as target GCPs. If the map content area is non-rectangular or the recognized coordinates fail to meet the heuristics, DIGMAPPER falls back to the place name extraction and the visual approach described below.

\noindent \textbf{Topographic Map Retrieval Using Place Names:} 
To automate title extraction from map images, DIGMAPPER leverages the GPT-4o\cite{gpt4o} multimodal vision-language model, prompting it to return only the map title from an input map image. The extracted title typically contains the quadrangle name, county, and state (e.g., \textit{Geologic Map of the Nazareth Quadrangle, Northampton County, Pennsylvania}). To improve retrieval precision, DIGMAPPER first applies GeoLM \cite{li2023geolm}, a geospatial domain-adapted language model fine-tuned for toponym recognition, to extract geographic entity names from the title. These recognized toponyms are then used in a fuzzy string-matching procedure to identify candidate matches from a curated database of approximately 250,000 georeferenced USGS topographic maps, as most geologic maps use the USGS topographic maps as their basemap. For georeferencing, DIGMAPPER aligns the detected corner points from Section \ref{subsec:georef-geocoord} to the bounding coordinates of the retrieved topographic maps, treating them respectively as source and target ground control points.
\sloppy
\subsubsection{Visual-based Method} \label{subsec:georef-viz}
Our visual-based georeferencing method matches image features between a query geologic map and candidate georeferenced USGS topographic maps. This method enhances fine-scale georeferencing through dense keypoint matching, especially when the text-based method is insufficient or fails.

First, DIGMAPPER generates a list of candidate topographic maps using the text-based methods described in \ref{subsec:georef-text}. It identifies either coordinates or bounding boxes from matching results and selects all topographic maps within a 10-kilometer buffer. This buffer accounts for possible misalignments during coordinate detection and was determined empirically. Next, DIGMAPPER applies a visual matcher using the LightGlue model \cite{lindenberger2023lightglue} for dense feature matching. A pretrained LightGlue model extracts keypoints from both the geologic and candidate topographic maps and computes potential keypoint pairs with associated confidence scores. These scores reflect the model’s certainty that each pair is a valid match. To estimate the spatial transformation, we apply RANSAC \cite{fischler1981random} to identify the homographic transformation that captures the most consistent keypoint matches. For each candidate, we compute the mean confidence score of the RANSAC inlier keypoints. The topographic map with the highest mean score is selected as the best match. If this mean score is below 0.5 (on a scale from 0 to 1), the match is considered low-confidence, and the visual pipeline discards it. To compute the final georeferencing transformation, we use only RANSAC inlier keypoint pairs with confidence scores above the 0.5 threshold. This transformation is represented by a 3×3 homography matrix, which maps pixel coordinates from the scanned geologic map to the corresponding positions on the topographic map.  Since computing a homography matrix requires at least four valid inlier keypoint pairs, the visual method cannot proceed if fewer than four are available. When four or more keypoints are found, the computed homography matrix can then be used to georeference the query geologic map.

\subsubsection{Evaluation}
To evaluate the performance of the georeferencing module, we tested 63 geologic maps in the DARPA-USGS dataset with two evaluation metrics: geodesic RMSE and pixel-normalized RMSE. The geodesic \cite{gritta2020pragmatic} RMSE (in kilometers) measures the root mean squared error between the predicted and ground-truth coordinates in geospatial space. The pixel-normalized RMSE computes the Euclidean error between predicted and ground-truth control points in pixel space, normalizes the error by the diagonal length of the image to ensure scale invariance, and then applies the RMSE formulation to quantify overall misalignment. 
We define the following thresholds to interpret RMSE values: $\text{RMSE}_{\text{geo}} < 0.1$ indicates negligible deviation from ground truth, $0.1 \leq \text{RMSE}_{\text{geo}} < 1$ reflects a noticeable but moderate shift, and $\text{RMSE}_{\text{geo}} \geq 1$ denotes a significant discrepancy. Similarly, for $\text{RMSE}_{\text{pixel-norm}}$, values below 0.1 correspond to visually imperceptible misalignments, suggesting excellent georeferencing results. 

\noindent\textbf{Text-based Results:}
\label{subsec:georef-text-results}
Across the 63 test maps, our method achieves a median geodesic RMSE ($\text{RMSE}_{\text{geo}}$) of 0.104 and a mean of 14.62, highlighting the robustness and accuracy of the text-based model. This low median error suggests strong map corner detection and geocoordinate recognition. A detailed breakdown of the georeferencing accuracy is provided in Table~\ref{tab:georef-text}. A $\text{RMSE}_{\text{pixel-norm}}$ below 0.1 suggests excellent alignment, indicating that minimal manual intervention is needed. In our evaluation, 34 out of 63 maps ($\approx$54\%) achieve pixel-scale RMSE values below this threshold. In general, the \textit{excellent} category contains maps with rectangular map content areas, clearly defined borders, and coordinate annotations at all four corners. The \textit{good} category includes maps that clearly specify the state, county, and quadrangle names, increasing the likelihood of alignment with existing topographic basemaps covering the same or adjacent regions. Maps classified as \textit{fair} often lack specific county or quadrangle names, instead providing only a general geographic description. For instance, a title such as ``Geologic Map of the Mountain Top Mercury Deposit, Southwestern Alaska'' may not correspond to any well-defined topographic basemap, thereby limiting its geospatial alignment potential.

\noindent\textbf {Visual-based Results:} Among the 32 maps rated as good or fair using the text-based method in Table~\ref{tab:georef-text}, the visual method enables to improve the results of nine maps, of which it obtains a mean $\text{RMSE}_{\text{geo}}$ of 2.10, a minimum value of 0.28 and a maximum value of 5.93. The visual method also includes a process of filtering out low-confidence results, ensuring that the visual georeferencing pipeline avoids outputting maps with extremely high RMSE errors. In particular, the maximum $\text{RMSE}_{\text{geo}}$ value, which, at 5.93, is much lower than the maximum $\text{RMSE}_{\text{geo}}$ in the text-based method for the same nine maps, which is 33.31. Similarly, the mean $\text{RMSE}_{\text{geo}}$ of the visual method is comparatively lower than the text-based mean $\text{RMSE}_{\text{geo}}$ of 6.04 over the set of nine maps. Using the same $\text{RMSE}_{\text{geo}}$ thresholds described in \ref{subsec:georef-text-results}, of the nine maps, six improve from a \textit{good} rating in the text pipeline to an \textit{excellent} rating with the visual pipeline. Two maps improve from \textit{fair} to \textit{excellent}, and one map improves from \textit{fair} to \textit{good}. In general, maps with \textit{excellent} georeferencing performance have distinct geographic features that are consistent between geologic and topographic maps, such as rivers, road lines, or coastlines. \textit{Good} Maps typically contain some shared geologic features but also exhibit significant visual differences between the geologic and topographic maps, such as distinct color schemes or varying contour line densities. Maps that fail georeferencing often face challenges: (1) incomplete overlap with any topographic basemap candidates, (2) significant scale discrepancies between the geologic and topographic maps, or (3) substantial visual occlusions on the geologic map. The challenges result in too few inlier point pairs, preventing the visual-based method from computing a valid homography matrix. However, when sufficient inlier point pairs are available to calculate the homography matrix, the visual-based method serves as a refinement step atop the text-based approach to improve the geolocation accuracy.

\begin{table}[ht]
\centering
\caption{Georeferencing quality breakdown of text-based approach based on $\text{RMSE}_{\text{geo}}$ thresholds}
\begin{tabular}{@{}lcc@{}}
\toprule
\textbf{Quality Category} & \textbf{$\text{RMSE}_{\text{geo}}$ Range} & \textbf{Map Count } \\
\midrule
Excellent         & $\text{RMSE}_{\text{geo}}$ $<$ 0.1   & 31   \\
Good   & $0.1 \leq$ $\text{RMSE}_{\text{geo}}$ $< 1$   &    8   \\
Fair   & $\text{RMSE}_{\text{geo}}$ $\geq$ 1    &    24   \\
\bottomrule
\end{tabular}
\label{tab:georef-text}
\end{table}
\vspace{-10pt}

\section{System Integration and Implementation}\label{sec:system}
\begin{figure}[htbp]
    \centering
    \includegraphics[width=\linewidth]{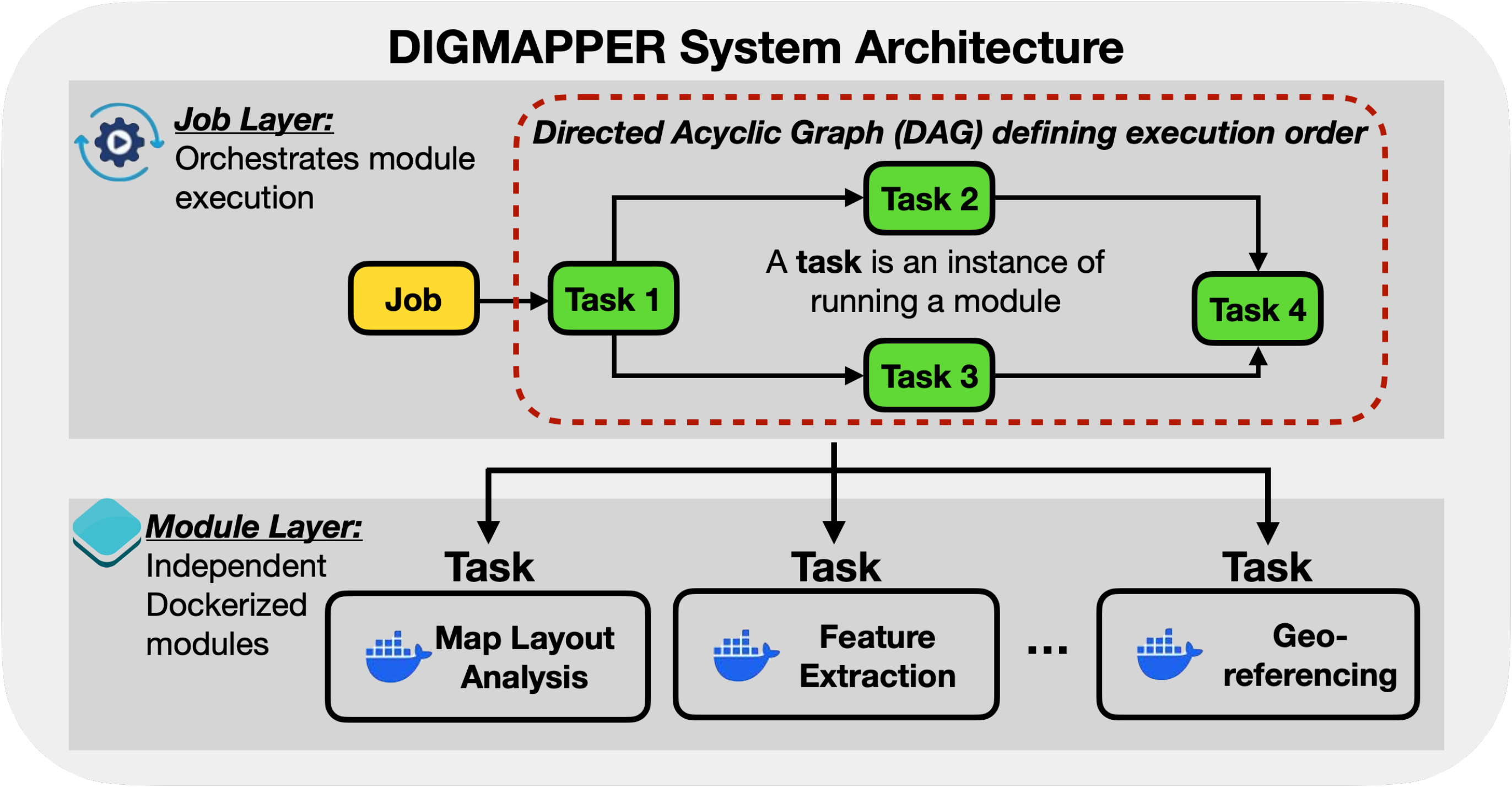}
    \caption{DIGMAPPER System Architecture consists of two layers. The Job Layer (top) orchestrates task scheduling using a Directed Acyclic Graph (DAG), where each task represents an instance of executing a module. The Module Layer (bottom) comprises independent Dockerized modules.}
    \label{fig:DIGMAPPER_arch}
\end{figure}

DIGMAPPER, illustrated in Figure~\ref{fig:DIGMAPPER_arch}, consists of two layers: the Module Layer and the Job Layer, which together support a scalable and flexible map digitization pipeline. At the Module Layer, the bottom part in Figure~\ref{fig:DIGMAPPER_arch}, each processing component, such as map layout analysis, is implemented as an independent Docker image. The use of Docker provides several key advantages: (1) portability across computing environments by packaging all dependencies, (2) isolation to prevent conflicts between modules, (3) standardized execution environments for reproducibility, and (4) support for independent development, testing, and updating of modules.

Above the Module Layer is the Job Layer (showing at the top of Figure~\ref{fig:DIGMAPPER_arch}), which orchestrates module execution using the Luigi Python library\footnote{\url{https://pypi.org/project/luigi/1.0.3/}}. A job is a digitization workflow composed of a sequence of tasks. For example, a job for point feature extraction involves tasks such as map layout analysis, map cropping, and point extraction. The task sequences are defined by a Directed Acyclic Graph (DAG), where each node represents a task (i.e., the execution of a specific module), and edges represent data dependencies between tasks. The DAG ensures correct execution order, enables fault tolerance by supporting task retries, and allows parallel execution of independent tasks. Additionally, DAG simplifies the integration of new modules by allowing users to specify upstream dependencies without altering the overall pipeline structure. In summary, the Docker-based modular design and DAG-based orchestration enable DIGMAPPER to scale efficiently to large map collections and quickly adapt to evolving digitization needs without requiring major system changes.

DIGMAPPER is deployed on an AWS g4dn.8xlarge instance with 32 vCPUs, 128 GiB of RAM, and a single NVIDIA T4 Tensor Core GPU with 16 GiB of GPU memory. Table~\ref{tab:module_timing} presents the average processing time, GPU usage per module, and total time per map. The most time-consuming modules are Polygon Extraction and Map Layout Analysis due to the high number of polygon features on geologic maps. Upgrading to a powerful GPU can reduce the processing time of the  Polygon Extraction module. Other modules, including Point and Line Extraction, run efficiently with GPU acceleration, while Georeferencing and Map Crop are lightweight. Due to DIGMAPPER’s modular architecture, modules execute in parallel, and the total runtime is bounded by the slowest module rather than the sum of all. Overall, DIGMAPPER processes a map in under 25 minutes, significantly faster than manual digitization, which typically takes hours per map.
\vspace{-5pt}
\begin{table}[htbp]
\small
\centering
\caption{Module-wise average processing time and GPU usage}
\label{tab:module_timing}
\begin{tabular}{l@{\hskip 8pt}c@{\hskip 8pt}c}
\hline
\textbf{Module} & \textbf{Avg. Time/Map (sec)} & \textbf{GPU Usage} \\
\hline
Map Layout Analysis& 542.14  & Yes \\
Map Crop           & 24.22  & No  \\
Polygon Extraction & 5596.54  & Yes \\
Line Extraction    & 337.49  & Yes \\
Point Extraction   & 68.51  & Yes \\
Georeferencing     & 105.44  & Yes  \\
\hline
\textbf{Total}     & 6162.90 & -- \\
\hline
\end{tabular}
\end{table}
\vspace{-10pt}
\section{Related Work}
\noindent
\textbf{Map Layout Analysis:}
Recent document layout analysis models~\cite{layoutlmv3,layoutparser,vsr_layout,udop_layout} are developed for general documents and do not incorporate the knowledge required to handle the unique structure and semantics of historical maps, such as geologic maps. For example, LayoutLMv3~\cite{layoutlmv3}, a transformer-based architecture that integrates text, layout, and visual information, segments distinct regions within documents, such as scientific papers and newspapers. Additionally, tools like LayoutParser\cite{layoutparser} provide a suite of pre-trained deep learning models, including Faster R-CNN and Detectron2-based detectors, which have been used to parse layout elements in scanned documents. However, these models are not trained on map images and thus lack the spatial and semantic understanding for cartographic content. Meanwhile, Large Language Models (LLMs) like GPT-4o~\cite{gpt4o} have demonstrated strong general knowledge across domains. In-context learning~\cite{incontext_survey}, where a model is guided by a few task-specific examples, has improved performance in various applications such as medical diagnosis~\cite{incontext_healthcare} and legal reasoning~\cite{incontext_legal}. This paper demonstrates that in-context learning can also be effectively applied to map layout analysis.

\noindent
\textbf{Polygon Extraction:} Previous research on polygon extraction can be categorized into single-feature, foreground-based, and multi-feature approaches. Single-feature methods focus on extracting polygons of a specific type with uniform color representation~\cite{arteaga2013historical,garcia2021potential,heitzler2020cartographic,wu2022leveraging,wu2023cross,xia2025mapsam}. These approaches do not dynamically incorporate legend items as input and are unsuitable for our task, which requires identifying diverse legend items with varying colors and markings across maps. Foreground-based methods, including foreground detection~\cite{arteaga2013historical,arzoumanidis2025semantic} and instance segmentation~\cite{wu2023cross,xia2025mapsam}, aim to delineate polygon boundaries using unsupervised~\cite{wang2022freesolo} or pre-trained models~\cite{kirillov2023segment,ravi2024sam}. These methods often struggle with closely nested boundaries or overlapping translucent symbols. Multi-feature approaches explicitly treat legend items as part of the input for convolutional models~\cite{lin2023exploiting,luo2023critical}. These methods can be generalized across a variety of legend items. However, their performance tends to degrade on low-resolution maps or unseen map styles due to a lack of training diversity or overfitting. As a multi-feature approach, TOPAZ in our system addresses the limitations by explicitly integrating heterogeneous visual and semantic cues from both legend items and map content within a unified encoder-decoder framework, enabling reliable polygon extraction across diverse map styles. 

\noindent
\textbf{Line Extraction:} SOTA line extraction methods are mainly categorized into segmentation-based and graph-based models. In the segmentation-based models, which classify pixels belonging to line features, CNN-based approaches introduce strip convolutional kernels~\cite{siinet,coanet,Fe-LinkNet,RADANet} and transformer-based models proposing axial self-attention mechanisms~\cite{segment_road_transformer:2023a} to capture directional image context. However, these methods often struggle with features that have arbitrary orientations, such as waterlines. To capture arbitrary orientations, researchers propose topology-aware loss functions~\cite{segment_road:2023b,segment_road:2023e}, feature refinement modules~\cite{segment_road:2023d}, and GAN-based approaches~\cite{segment_road_gan:2023c,segment_road_gan:2023f}. However, due to independent pixel predictions, segmentation-based methods may predict fragmented lines by missing a few pixels along the path. In contrast, graph-based models directly construct graphs from images, using RNNs~\cite{polymapper} or transformers~\cite{road_sam:2024,transformer:roads5,relationformer,segment_road_ternformer:2023,TD-Road,road-topo}. For example, RNGDet~\cite{transformer:roads5} iteratively predicts adjacent nodes but relies on limited context, whereas Relationformer~\cite{relationformer} enhances context modeling by incorporating all nodes. Nonetheless, Relationformer still struggles with explicitly modeling connectivity relationships, resulting in inaccuracies in graph generation. To capture sufficient connectivity relationships, LDTR in our system, built on Relationformer~\cite{relationformer}, utilizes a novel N-hop connectivity prediction module, which enables each node to aggregate spatial context from both adjacent and N-hop distant nodes. The sufficient spatial context allows LDTR to capture line orientations, curvatures, and topological structures. Consequently, LDTR achieves accurate predictions of line connectivity.  

\noindent
\textbf{Point Extraction:} 
Existing work on extracting point symbols can be generally categorized into template matching~\cite{opencv_library} and object detection~\cite{redmon2016lookonceunifiedrealtime,girshick2015fastrcnn,yolov8}. Template matching is a traditional method that matches preprocessed map features with some predefined templates~\cite{7575646,budig2015active}. However, these methods have shown limited performance compared to modern object detection methods, primarily due to their sensitivity to variations in symbol scales and orientations, as well as their limitations in handling diverse map qualities and styles. In contrast, applying object detection techniques to point symbol detection~\cite{Huang2023PointSymbol,smith2025estimating} has led to significant performance improvements, although these methods typically require a large number of labeled instances for training. DIGMAPPER's point symbol detection module utilizes an object detection model trained on both synthetic and human-labeled data, thereby mitigating the limited availability of human annotations. 

\noindent
\textbf{Georeferencing:} Previous work has employed corner point detection in conjunction with map sheet ID recognition to facilitate georeferencing, as demonstrated by Gede et al. \cite{gede2021automatic}. Building on this, our approach similarly incorporates corner detection but integrates it with the extraction of geographic coordinates. More broadly, research on automatic georeferencing often leverages visual patterns commonly found in geologic, historical, and topographic maps—such as lines, symbols, and other graphical elements. For instance, several studies focus on detecting line features, such as road networks, in both orthoimagery and scanned maps, underscoring the importance of visual cues in the georeferencing process \cite{chen2008automatically, chiang2009automatic}. Other methods adopt a more general strategy by aligning vector representations of map content—such as quadrangle boundaries, roads, or natural features like rivers—between query maps and georeferenced references \cite{duan2017automatic, howe2019deformable, luft2021automatic}. These approaches typically assume some prior knowledge of the symbols likely to appear on a given map and their characteristic visual forms. In contrast, DIGMAPPER's text-based and visual-based georeferencing methods do not assume a specific type of matching geographic features and can accommodate a wide range of maps.
\vspace{-1pt}
\section{Conclusion}
This paper presents DIGMAPPER, a modular and scalable system for the automatic digitization of geologic maps. DIGMAPPER has already processed hundreds of maps to support real-world workflows for identifying potential locations of critical minerals such as nickel, lithium, rare earth elements, zinc, and porphyry copper, and has been successfully transitioned to the USGS to support critical mineral assessments. DIGMAPPER’s innovative modules address the limitations of training data and complex map contexts to accurately digitize geologic maps. DIGMAPPER's system design includes Dockerized modules for independent development and a DAG-based job orchestration layer that supports parallel execution and fault tolerance. Experimental results show that DIGMAPPER processes a single map under 25 minutes, significantly faster than manual digitization, which typically takes several hours. Overall, DIGMAPPER provides a practical and adaptable solution for converting large volumes of historical maps into analysis-ready geospatial data. Future work will extend DIGMAPPER to support additional map types by leveraging one-shot and few-shot learning techniques, reducing reliance on large annotated datasets.

\section{Acknowledgments}
This material is based upon works supported by the Defense Advanced Research Projects Agency (DARPA) under Agreement No. HR00112390132 and Contract No. 140D0423C0093. We thank Dr. Graham W. Lederer (USGS), Margaret A. Goldman (USGS), and David Watkins (USGS) for their help and collaboration on this project.

\newpage
\bibliographystyle{abbrv}
\bibliography{sigspatial2025}

\begin{thebibliography}{10}

\bibitem{segment_road:2023e}
A.~Abdollahi, B.~Pradhan, and A.~Alamri.
\newblock Sc-roaddeepnet: A new shape and connectivity-preserving road extraction deep learning-based network from remote sensing data.
\newblock {\em IEEE Transactions on Geoscience and Remote Sensing}, 60:1--15, 2022.

\bibitem{arteaga2013historical}
M.~G. Arteaga.
\newblock Historical map polygon and feature extractor.
\newblock In {\em Proceedings of the 1st ACM SIGSPATIAL International Workshop on MapInteraction}, pages 66--71, 2013.

\bibitem{arzoumanidis2025semantic}
L.~Arzoumanidis, J.~Knechtel, J.-H. Haunert, and Y.~Dehbi.
\newblock Semantic segmentation of historical maps using self-constructing graph convolutional networks.
\newblock {\em Cartography and Geographic Information Science}, pages 1--11, 2025.

\bibitem{budig2015active}
B.~Budig and T.~C. van Dijk.
\newblock Active learning for classifying template matches in historical maps.
\newblock In {\em Discovery Science: 18th International Conference, DS 2015, Banff, AB, Canada, October 4-6, 2015. Proceedings 18}, pages 33--47. Springer, 2015.

\bibitem{chen2008automatically}
C.-C. Chen, C.~A. Knoblock, and C.~Shahabi.
\newblock Automatically and accurately conflating raster maps with orthoimagery.
\newblock {\em GeoInformatica}, 12:377--410, 2008.

\bibitem{10.5555/3383708}
Y.-Y. Chiang, W.~Duan, S.~Leyk, J.~H. Uhl, and C.~A. Knoblock.
\newblock {\em Using Historical Maps in Scientific Studies: Applications, Challenges, and Best Practices}.
\newblock Springer Publishing Company, Incorporated, 1st edition, 2019.

\bibitem{chiang2009automatic}
Y.-Y. Chiang, C.~A. Knoblock, C.~Shahabi, and C.-C. Chen.
\newblock Automatic and accurate extraction of road intersections from raster maps.
\newblock {\em GeoInformatica}, 13:121--157, 2009.

\bibitem{10.1145/2557423}
Y.-Y. Chiang, S.~Leyk, and C.~A. Knoblock.
\newblock A survey of digital map processing techniques.
\newblock {\em ACM Comput. Surv.}, 47(1), May 2014.

\bibitem{fault_importance}
B.~Dadi, M.~Ouchchen, F.~Faik, S.~Boutaleb, D.~El~Azzab, Y.~Mamouch, L.~Achkouch, A.~Bajadi, E.~H. Abia, and B.~Sadeghi.
\newblock Hydrothermal fluid pathways and mineralization potential in the high atlas massif (morocco) using fuzzy logic and multifractal modeling.
\newblock {\em Ore Geology Reviews}, page 106401, 2024.

\bibitem{RADANet}
L.~Dai, G.~Zhang, and R.~Zhang.
\newblock Radanet: Road augmented deformable attention network for road extraction from complex high-resolution remote-sensing images.
\newblock {\em IEEE Trans. Geosci. Remote Sens.}, 2023.

\bibitem{incontext_survey}
Q.~Dong, L.~Li, D.~Dai, C.~Zheng, J.~Ma, R.~Li, H.~Xia, J.~Xu, Z.~Wu, T.~Liu, et~al.
\newblock A survey on in-context learning.
\newblock {\em arXiv preprint arXiv:2301.00234}, 2022.

\bibitem{map_challenge_1}
W.~Duan, Y.~Chiang, C.~A. Knoblock, S.~Leyk, and J.~Uhl.
\newblock Automatic generation of precisely delineated geographic features from georeferenced historical maps using deep learning.
\newblock In {\em Proceedings of the AutoCarto}, 2018.

\bibitem{duan2017automatic}
W.~Duan, Y.-Y. Chiang, C.~A. Knoblock, V.~Jain, D.~Feldman, J.~H. Uhl, and S.~Leyk.
\newblock Automatic alignment of geographic features in contemporary vector data and historical maps.
\newblock In {\em Proceedings of the 1st workshop on artificial intelligence and deep learning for geographic knowledge discovery}, pages 45--54, 2017.

\bibitem{cnn_railroads}
W.~Duan, Y.-Y. Chiang, S.~Leyk, J.~Uhl, and C.~Knoblock.
\newblock A label correction algorithm using prior information for automatic and accurate geospatial object recognition.
\newblock In {\em IEEE Big Data}, pages 1604--1610. IEEE, 2021.

\bibitem{label_correction}
W.~Duan, Y.-Y. Chiang, S.~Leyk, J.~H. Uhl, and C.~A. Knoblock.
\newblock A label correction algorithm using prior information for automatic and accurate geospatial object recognition.
\newblock In {\em 2021 IEEE International Conference on Big Data (Big Data)}, pages 1604--1610. IEEE, 2021.

\bibitem{review_ml_mine}
N.~Dumakor-Dupey and S.~Arya.
\newblock Machine learning—a review of applications in mineral resource estimation.
\newblock {\em Energies}, 14(14):4079, 2021.

\bibitem{arcgis_georeferencing}
{Esri}.
\newblock Georeferencing a raster automatically.
\newblock Accessed: 2025-05-07.

\bibitem{fischler1981random}
M.~A. Fischler and R.~C. Bolles.
\newblock Random sample consensus: a paradigm for model fitting with applications to image analysis and automated cartography.
\newblock {\em Communications of the ACM}, 24(6):381--395, 1981.

\bibitem{garcia2021potential}
A.~Garcia-Molsosa, H.~A. Orengo, D.~Lawrence, G.~Philip, K.~Hopper, and C.~A. Petrie.
\newblock Potential of deep learning segmentation for the extraction of archaeological features from historical map series.
\newblock {\em Archaeological Prospection}, 28(2):187--199, 2021.

\bibitem{gede2021automatic}
M.~Gede and L.~Varga.
\newblock Automatic georeferencing of topographic map sheets using opencv and tesseract.
\newblock In {\em Proceedings of the ICA}, volume~4, pages 1--4. Copernicus GmbH, 2021.

\bibitem{girshick2015fastrcnn}
R.~Girshick.
\newblock Fast r-cnn, 2015.

\bibitem{gritta2020pragmatic}
M.~Gritta, M.~T. Pilehvar, and N.~Collier.
\newblock A pragmatic guide to geoparsing evaluation: Toponyms, named entity recognition and pragmatics.
\newblock {\em Language resources and evaluation}, 54:683--712, 2020.

\bibitem{segment_road:2023b}
L.~Han, L.~Hou, X.~Zheng, Z.~Ding, H.~Yang, and K.~Zheng.
\newblock Segmentation is not the end of road extraction: An all-visible denoising autoencoder for connected and smooth road reconstruction.
\newblock {\em IEEE Transactions on Geoscience and Remote Sensing}, 61:1--18, 2023.

\bibitem{TD-Road}
Y.~He, R.~Garg, and A.~R. Chowdhury.
\newblock Td-road: Top-down road network extraction with holistic graph construction.
\newblock In {\em ECCV 2022}, pages 562--577. Springer, 2022.

\bibitem{correct_complete}
C.~Heipke, H.~Mayer, C.~Wiedemann, and O.~Jamet.
\newblock Evaluation of automatic road extraction.
\newblock {\em IAPRS}, 32(3 SECT 4W2):151--160, 1997.

\bibitem{heitzler2020cartographic}
M.~Heitzler and L.~Hurni.
\newblock Cartographic reconstruction of building footprints from historical maps: A study on the swiss siegfried map.
\newblock {\em Transactions in GIS}, 24(2):442--461, 2020.

\bibitem{road_sam:2024}
C.~Hetang, H.~Xue, C.~Le, T.~Yue, W.~Wang, and Y.~He.
\newblock Segment anything model for road network graph extraction.
\newblock In {\em Proceedings of the IEEE/CVF Conference on Computer Vision and Pattern Recognition}, pages 2556--2566, 2024.

\bibitem{howe2019deformable}
N.~R. Howe, J.~Weinman, J.~Gouwar, and A.~Shamji.
\newblock Deformable part models for automatically georeferencing historical map images.
\newblock In {\em Proceedings of the 27th ACM SIGSPATIAL International Conference on Advances in Geographic Information Systems}, pages 540--543, 2019.

\bibitem{future_ml_mine}
J.~Hronsky and O.~Kreuzer.
\newblock Applying spatial prospectivity mapping to exploration targeting: Fundamental practical issues and suggested solutions for the future.
\newblock {\em Ore Geology Reviews}, 107:647--653, 2019.

\bibitem{Huang2023PointSymbol}
W.~Huang, Q.~Sun, A.~Yu, W.~Guo, Q.~Xu, B.~Wen, and L.~Xu.
\newblock Leveraging deep convolutional neural network for point symbol recognition in scanned topographic maps.
\newblock {\em ISPRS International Journal of Geo-Information}, 12(3):128, 2023.

\bibitem{layoutlmv3}
Y.~Huang, T.~Lv, L.~Cui, Y.~Lu, and F.~Wei.
\newblock Layoutlmv3: Pre-training for document ai with unified text and image masking.
\newblock In {\em Proceedings of the 30th ACM international conference on multimedia}, pages 4083--4091, 2022.

\bibitem{opencv_library}
Itseez.
\newblock Opencv: Open source computer vision library.
\newblock In {\em OpenCV Library}, 2015.

\bibitem{yolov8}
G.~Jocher, A.~Chaurasia, Laughing-Q, J.~Fang, and A.~V.
\newblock Yolov8 - ultralytics.
\newblock \url{https://github.com/ultralytics/ultralytics}, 2023.
\newblock Accessed: 2025-05-11.

\bibitem{10.1145/3589132.3625579}
J.~Kim, Z.~Li, Y.~Lin, M.~Namgung, L.~Jang, and Y.-Y. Chiang.
\newblock The mapkurator system: A complete pipeline for extracting and linking text from historical maps.
\newblock In {\em Proceedings of the 31st ACM International Conference on Advances in Geographic Information Systems}, SIGSPATIAL '23, New York, NY, USA, 2023. Association for Computing Machinery.

\bibitem{kirillov2023segment}
A.~Kirillov, E.~Mintun, N.~Ravi, H.~Mao, C.~Rolland, L.~Gustafson, T.~Xiao, S.~Whitehead, A.~C. Berg, W.-Y. Lo, et~al.
\newblock Segment anything.
\newblock {\em arXiv preprint arXiv:2304.02643}, 2023.

\bibitem{lewis2019bart}
M.~Lewis, Y.~Liu, N.~Goyal, M.~Ghazvininejad, A.~Mohamed, O.~Levy, V.~Stoyanov, and L.~Zettlemoyer.
\newblock Bart: Denoising sequence-to-sequence pre-training for natural language generation, translation, and comprehension.
\newblock {\em arXiv preprint arXiv:1910.13461}, 2019.

\bibitem{segment_road_gan:2023f}
T.~Li, R.~Li, S.~Ye, Z.~Zhang, Z.~Yin, S.~Li, and Z.~Pan.
\newblock Crtgan: Controllable road network graphs generation via transformer based gan.
\newblock In {\em 2024 International Joint Conference on Neural Networks (IJCNN)}, pages 1--8. IEEE, 2024.

\bibitem{li2020automatic}
Z.~Li, Y.-Y. Chiang, S.~Tavakkol, B.~Shbita, J.~H. Uhl, S.~Leyk, and C.~A. Knoblock.
\newblock An automatic approach for generating rich, linked geo-metadata from historical map images.
\newblock In {\em Proceedings of the 26th ACM SIGKDD International Conference on Knowledge Discovery \& Data Mining}, pages 3290--3298, 2020.

\bibitem{polymapper}
Z.~Li, J.~D. Wegner, and A.~Lucchi.
\newblock Topological map extraction from overhead images.
\newblock In {\em ICCV}, pages 1715--1724, 2019.

\bibitem{li2023geolm}
Z.~Li, W.~Zhou, Y.-Y. Chiang, and M.~Chen.
\newblock Geolm: Empowering language models for geospatially grounded language understanding.
\newblock {\em arXiv preprint arXiv:2310.14478}, 2023.

\bibitem{lin2023exploiting}
F.~Lin, C.~A. Knoblock, B.~Shbita, B.~Vu, Z.~Li, and Y.-Y. Chiang.
\newblock Exploiting polygon metadata to understand raster maps-accurate polygonal feature extraction.
\newblock In {\em Proceedings of the 31st ACM International Conference on Advances in Geographic Information Systems}, pages 1--12, 2023.

\bibitem{segment_road_gan:2023c}
S.~Lin, X.~Yao, X.~Liu, S.~Wang, H.-M. Chen, L.~Ding, J.~Zhang, G.~Chen, and Q.~Mei.
\newblock Ms-agan: Road extraction via multi-scale information fusion and asymmetric generative adversarial networks from high-resolution remote sensing images under complex backgrounds.
\newblock {\em Remote Sensing}, 15(13):3367, 2023.

\bibitem{lin2024hyper}
Y.~Lin and Y.-Y. Chiang.
\newblock Hyper-local deformable transformers for text spotting on historical maps.
\newblock In {\em Proceedings of the 30th ACM SIGKDD Conference on Knowledge Discovery and Data Mining}, pages 5387--5397, 2024.

\bibitem{lindenberger2023lightglue}
P.~Lindenberger, P.-E. Sarlin, and M.~Pollefeys.
\newblock Lightglue: Local feature matching at light speed.
\newblock In {\em Proceedings of the IEEE/CVF International Conference on Computer Vision}, pages 17627--17638, 2023.

\bibitem{luft2021automatic}
J.~Luft and J.~Schiewe.
\newblock Automatic content-based georeferencing of historical topographic maps.
\newblock {\em Transactions in GIS}, 25(6):2888--2906, 2021.

\bibitem{luo2023critical}
S.~Luo, A.~Saxton, A.~Bode, P.~Mazumdar, and V.~Kindratenko.
\newblock Critical minerals map feature extraction using deep learning.
\newblock {\em IEEE Geoscience and Remote Sensing Letters}, 20:1--5, 2023.

\bibitem{coanet}
J.~Mei, R.-J. Li, W.~Gao, and M.-M. Cheng.
\newblock Coanet: Connectivity attention network for road extraction from satellite imagery.
\newblock {\em TIP}, 30:8540--8552, 2021.

\bibitem{7575646}
Q.~Miao, P.~Xu, X.~Li, J.~Song, W.~Li, and Y.~Yang.
\newblock The recognition of the point symbols in the scanned topographic maps.
\newblock {\em IEEE Transactions on Image Processing}, 26(6):2751--2766, 2017.

\bibitem{trajectory_refinement}
M.~Musleh and M.~Mokbel.
\newblock A demonstration of kamel: A scalable bert-based system for trajectory imputation.
\newblock In {\em Companion of the 2023 International Conference on Management of Data}, pages 191--194, 2023.

\bibitem{incontext_healthcare}
F.~Nazary, Y.~Deldjoo, T.~Di~Noia, and E.~Di~Sciascio.
\newblock Xai4llm. let machine learning models and llms collaborate for enhanced in-context learning in healthcare.
\newblock {\em arXiv preprint arXiv:2405.06270}, 2024.

\bibitem{gpt4o}
OpenAI.
\newblock Gpt-4o technical report.
\newblock \url{https://openai.com/research/gpt-4o}, 2024.
\newblock Accessed: YYYY-MM-DD.

\bibitem{segment_road:2023d}
L.~Qiu, D.~Yu, C.~Zhang, and X.~Zhang.
\newblock A semantics-geometry framework for road extraction from remote sensing images.
\newblock {\em IEEE Geoscience and Remote Sensing Letters}, 20:1--5, 2023.

\bibitem{ravi2024sam}
N.~Ravi, V.~Gabeur, Y.-T. Hu, R.~Hu, C.~Ryali, T.~Ma, H.~Khedr, R.~R{\"a}dle, C.~Rolland, L.~Gustafson, et~al.
\newblock Sam 2: Segment anything in images and videos.
\newblock {\em arXiv preprint arXiv:2408.00714}, 2024.

\bibitem{redmon2016lookonceunifiedrealtime}
J.~Redmon, S.~Divvala, R.~Girshick, and A.~Farhadi.
\newblock You only look once: Unified, real-time object detection, 2016.

\bibitem{layoutparser}
Z.~Shen, R.~Zhang, M.~Dell, B.~C.~G. Lee, J.~Carlson, and W.~Li.
\newblock Layoutparser: A unified toolkit for deep learning based document image analysis.
\newblock In {\em Document Analysis and Recognition--ICDAR 2021: 16th International Conference, Lausanne, Switzerland, September 5--10, 2021, Proceedings, Part I 16}, pages 131--146. Springer, 2021.

\bibitem{relationformer}
S.~Shit, R.~Koner, B.~Wittmann, J.~Paetzold, I.~Ezhov, H.~Li, J.~Pan, S.~Sharifzadeh, G.~Kaissis, V.~Tresp, et~al.
\newblock Relationformer: A unified framework for image-to-graph generation.
\newblock In {\em ECCV}, pages 422--439. Springer, 2022.

\bibitem{smith2025estimating}
E.~S. Smith, C.~Fleet, S.~King, W.~Mackaness, H.~Walker, and C.~E. Scott.
\newblock Estimating the density of urban trees in 1890s leeds and edinburgh using object detection on historical maps.
\newblock {\em Computers, Environment and Urban Systems}, 115:102219, 2025.

\bibitem{subcommittee2006fgdc}
G.~D. Subcommittee.
\newblock Fgdc digital cartographic standard for geologic map symbolization.
\newblock Technical report, Citeseer, 2006.

\bibitem{udop_layout}
Z.~Tang, Z.~Yang, G.~Wang, Y.~Fang, Y.~Liu, C.~Zhu, M.~Zeng, C.~Zhang, and M.~Bansal.
\newblock Unifying vision, text, and layout for universal document processing.
\newblock In {\em Proceedings of the IEEE/CVF conference on computer vision and pattern recognition}, pages 19254--19264, 2023.

\bibitem{siinet}
C.~Tao, J.~Qi, Y.~Li, H.~Wang, and H.~Li.
\newblock Spatial information inference net: Road extraction using road-specific contextual information.
\newblock {\em IJPRS}, 158:155--166, 2019.

\bibitem{segment_road_ternformer:2023}
B.~Wang, Q.~Liu, Z.~Hu, W.~Wang, and Y.~Wang.
\newblock Ternformer: Topology-enhanced road network extraction by exploring local connectivity.
\newblock {\em IEEE Transactions on Geoscience and Remote Sensing}, 2023.

\bibitem{segment_road_transformer:2023a}
C.~Wang, R.~Xu, S.~Xu, W.~Meng, R.~Wang, J.~Zhang, and X.~Zhang.
\newblock Toward accurate and efficient road extraction by leveraging the characteristics of road shapes.
\newblock {\em IEEE Transactions on Geoscience and Remote Sensing}, 61:1--16, 2023.

\bibitem{Fe-LinkNet}
Q.~Wang, H.~Bai, C.~He, and J.~Cheng.
\newblock Fe-linknet: Enhanced d-linknet with attention and dense connection for road extraction in high-resolution remote sensing images.
\newblock In {\em IGARSS}, pages 3043--3046. IEEE, 2022.

\bibitem{wang2022freesolo}
X.~Wang, Z.~Yu, S.~De~Mello, J.~Kautz, A.~Anandkumar, C.~Shen, and J.~M. Alvarez.
\newblock Freesolo: Learning to segment objects without annotations.
\newblock In {\em Proceedings of the IEEE/CVF Conference on Computer Vision and Pattern Recognition}, pages 14176--14186, 2022.

\bibitem{ai_mine}
J.~Woodhead and M.~Landry.
\newblock Harnessing the power of artificial intelligence and machine learning in mineral exploration—opportunities and cautionary notes.
\newblock {\em SEG Discovery}, (127):19--31, 2021.

\bibitem{wu2023cross}
S.~Wu, Y.~Chen, K.~Schindler, and L.~Hurni.
\newblock Cross-attention spatio-temporal context transformer for semantic segmentation of historical maps.
\newblock In {\em Proceedings of the 31st ACM International Conference on Advances in Geographic Information Systems}, pages 1--9, 2023.

\bibitem{wu2022leveraging}
S.~Wu, M.~Heitzler, and L.~Hurni.
\newblock Leveraging uncertainty estimation and spatial pyramid pooling for extracting hydrological features from scanned historical topographic maps.
\newblock {\em GIScience \& Remote Sensing}, 59(1):200--214, 2022.

\bibitem{xia2025mapsam}
X.~Xia, D.~Zhang, W.~Song, W.~Huang, and L.~Hurni.
\newblock Mapsam: adapting segment anything model for automated feature detection in historical maps.
\newblock {\em GIScience \& Remote Sensing}, 62(1):2494883, 2025.

\bibitem{transformer:roads5}
Z.~Xu, Y.~Liu, L.~Gan, Y.~Sun, X.~Wu, M.~Liu, and L.~Wang.
\newblock Rngdet: Road network graph detection by transformer in aerial images.
\newblock {\em IEEE Trans. Geosci. Remote Sens.}, 60:1--12, 2022.

\bibitem{incontext_legal}
R.~Yao, Y.~Wu, C.~Wang, J.~Xiong, F.~Wang, and X.~Liu.
\newblock Elevating legal llm responses: Harnessing trainable logical structures and semantic knowledge with legal reasoning.
\newblock {\em arXiv preprint arXiv:2502.07912}, 2025.

\bibitem{ml_mine}
C.~Yeomans, R.~Shail, S.~Grebby, V.~Nyk{\"a}nen, M.~Middleton, and P.~Lusty.
\newblock A machine learning approach to tungsten prospectivity modelling using knowledge-driven feature extraction and model confidence.
\newblock {\em Geoscience Frontiers}, 11(6):2067--2081, 2020.

\bibitem{road-topo}
J.~Zhang, X.~Hu, Y.~Wei, and L.~Zhang.
\newblock Road topology extraction from satellite imagery by joint learning of nodes and their connectivity.
\newblock {\em IEEE Trans. Geosci. Remote Sens.}, 61:1--13, 2023.

\bibitem{vsr_layout}
P.~Zhang, C.~Li, L.~Qiao, Z.~Cheng, S.~Pu, Y.~Niu, and F.~Wu.
\newblock Vsr: a unified framework for document layout analysis combining vision, semantics and relations.
\newblock In {\em Document Analysis and Recognition--ICDAR 2021: 16th International Conference, Lausanne, Switzerland, September 5--10, 2021, Proceedings, Part I 16}, pages 115--130. Springer, 2021.

\bibitem{deform-detr}
X.~Zhu, W.~Su, L.~Lu, B.~Li, X.~Wang, and J.~Dai.
\newblock Deformable detr: Deformable transformers for end-to-end object detection.
\newblock {\em arXiv preprint arXiv:2010.04159}, 2020.

\end{thebibliography}

\end{document}